\documentclass{bmvc2k}

\usepackage{times}
\usepackage{epsfig}
\usepackage{graphicx}
\usepackage{amsmath}
\usepackage{amssymb}
\usepackage{kotex}      % Korean
\usepackage{xcolor}     % 
\usepackage{multirow, makecell} % Table
\usepackage{booktabs}   % Table (\toprule, \cmidrule)
\usepackage{graphicx,subfigure}
\usepackage{wrapfig}
\newcommand{\bh}{\mathbf{h}}

\newcommand{\bk}{\mathbf{k}}

\newcommand{\bu}{\mathbf{u}}
\newcommand{\bv}{\mathbf{v}}

\newcommand{\bz}{\mathbf{z}}

\newcommand{\bM}{\mathbf{M}}

\newcommand{\calH}{{\mathcal{H}}}

\newcommand{\calL}{{\mathcal{L}}}
\newcommand{\calM}{{\mathcal{M}}}
\newcommand{\calN}{{\mathcal{N}}}

\newcommand{\calS}{{\mathcal{S}}}
\newcommand{\calT}{{\mathcal{T}}}

\newcommand{\calZ}{{\mathcal{Z}}}

\newcommand{\bbE}{\mathbb{E}}

\newcommand{\bbR}{\mathbb{R}}

\def\[#1\]{\begin{align}#1\end{align}}
\title{Continuous-Time Video Generation via \\ Learning Motion Dynamics with Neural ODE}

\addauthor{Kangyeol Kim*}{kangyeolk@kaist.ac.kr}{1,7}
\addauthor{Sunghyun Park*$^{\dagger}$}{psh01087@kaist.ac.kr}{1,2}
\addauthor{Junsoo Lee}{junsoolee93@webtoonscorp.com}{3}
\addauthor{Joonseok Lee}{joonseok2010@gmail.com}{4,5}
\addauthor{Sookyung Kim}{sookim@parc.com}{6}
\addauthor{Jaegul Choo}{jchoo@kaist.ac.kr}{1,7}
\addauthor{Edward Choi}{edwardchoi@kaist.ac.kr}{1}

\addinstitution{KAIST AI \\ Korea}
\addinstitution{Kakao Enterprise \\ Korea}
\addinstitution{Naver Webtoon \\ Korea}
\addinstitution{Seoul National University \\ Korea}
\addinstitution{Google Research \\ United States}
\addinstitution{Xerox PARC \\ United States}
\addinstitution{Letsur Inc. \\ Korea}

\runninghead{Kim et al}{Continuous-Time Video Generation}

\def\eg{\emph{e.g}\bmvaOneDot}

\begin{document}

%%%%%%%%% Additional command
\newcommand{\model}{MODE-GAN\xspace}
\newcommand{\fusion}{composition block\xspace}

\newcommand{\ecedit}[1]{\textcolor{blue}{\emph{[ED: #1]}}}
\newcommand{\sooedit}[1]{\textcolor{green}{\emph{[Soo: #1]}}}
\newcommand{\jsedit}[1]{\textcolor{orange}{\emph{[JS: #1]}}}
\newcommand{\jg}[1]{\todored{JG: #1}}
\newcommand{\js}[1]{\todoblue{JS: #1}}

\newcommand{\msedit}[1]{\textcolor{red}{\emph{[MS: #1]}}}
\newcommand{\kyedit}[1]{\textcolor{purple}{\emph{[KY: #1]}}}
\newcommand{\shedit}[1]{\textcolor{brown}{\emph{[SH: #1]}}}

\newcommand{\Romannum}[1]{I}
\newcommand{\Romannumm}[1]{II}

\definecolor{blackpink}{rgb}{0.6,0,0.6}
\definecolor{mint}{RGB}{0, 220, 190}

\newcommand{\rnum}[1]{{\color{mint}{#1 }}}
\newcommand{\rev}[1]{{\color{blackpink}{#1}}}
\newcommand{\titlesize}{\fontsize{10}{11pt}\selectfont}

%%%%%%%%% TITLE
\maketitle

%%%%%%%%% ABSTRACT
\vspace{-1mm}
\begin{abstract}
\vspace{-1mm}
In order to perform unconditional video generation, we must learn the distribution of the real-world videos.
In an effort to synthesize high-quality videos, various studies attempted to learn a mapping function between noise and videos, including recent efforts to separate motion distribution and appearance distribution.
Previous methods, however, learn motion dynamics in discretized, fixed-interval timesteps, which is contrary to the continuous nature of motion of a physical body.
In this paper, we propose a novel video generation approach that learns separate distributions for motion and appearance, the former modeled by neural Ordinary Differential Equation (ODE) to learn natural motion dynamics.
Specifically, we employ a two-stage approach where the first stage converts a noise vector to a sequence of keypoints in arbitrary frame rates, and the second stage synthesizes videos based on the given keypoints sequence and the appearance noise vector.
Our model not only quantitatively outperforms recent baselines for video generation, but also demonstrates versatile functionality such as dynamic frame rate manipulation and motion transfer between two datasets, thus opening new doors to diverse video generation applications.
\end{abstract}

%%%%%%%%% BODY TEXT
\vspace{-0.1cm}
\section{Introduction}
\label{sec:intro}
\vspace{-0.1cm}

Creating realistic videos from scratch (\textit{i.e.}, unconditional video generation (UVG)) requires the model to learn the distribution of the video data.
In other words, we are essentially learning a density function from which we can sample unseen videos.
This is typically achieved by employing an adversarial training framework, thanks to the advances in deep generative models where GAN-based approaches have shown impressive performance in static image generation ~\cite{karras2019style,karras2020analyzing,cho2019image,lee2020drit++}.
Video generation, however, differs from image generation since we must consider the temporal aspect in addition to the spatial aspect.

Initial UVG approaches tried to learn a single density function for both temporal and spatial aspects, where based on a single noise vector, a sequence of video frames were generated at each timestep~\cite{saito2017temporal,vondrick2016generating}. 
Such approaches, however, were limited to generating short videos with simple dynamics (\textit{e.g.}, linear motions such as moving trains).
This limitation stems most likely from failing to treat the two aspects effectively.
Specifically, a video can be decomposed into two orthogonal elements: the motion (\textit{i.e.}, movement of an object(s)) and the appearance (\textit{i.e.}, background, style of the object(s), etc.), where the former is concerned with the temporal aspect, and the latter the spatial aspect.

\begin{figure}[t!]
    \centering
    \includegraphics[width=0.7\linewidth]{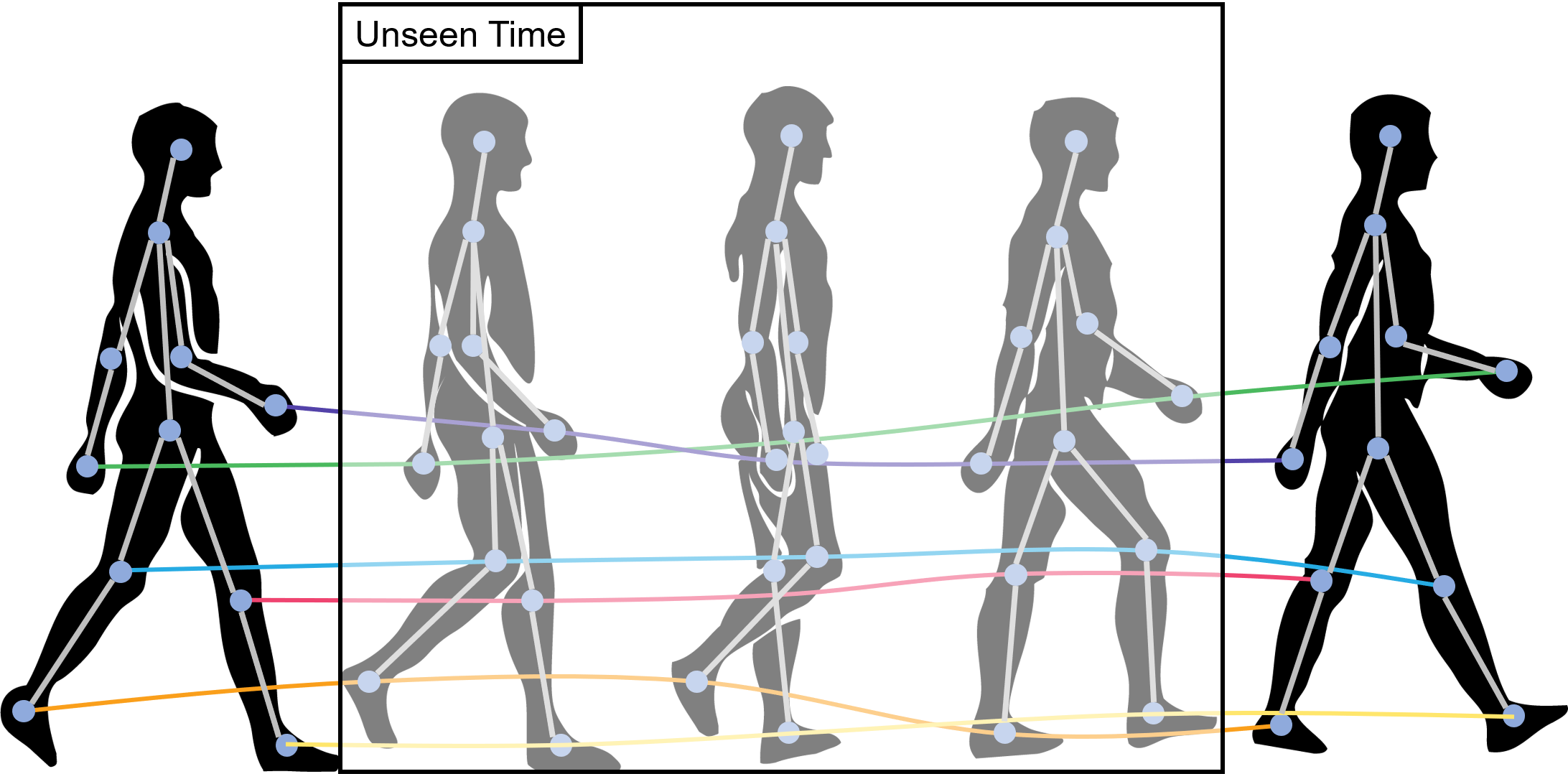}
    \vspace{-0.3cm}
    \caption{A walking human and corresponding keypoints. Rigid treatment of time as discretized, fixed-interval timesteps (\textit{the 1$^{st}$ and 5$^{th}$ steps}) prevents the model from learning the underlying motion dynamics by missing out frames at unseen timesteps (\textit{motion in the box}).}
    \vspace{-0.6cm}
    \label{fig:motivation}
\end{figure}

Based on this observation, a couple of recent works tried to learn separate distributions (one for motion and another for appearance) and have shown improved video quality as well as more natural motion dynamics~\cite{tulyakov2018mocogan,wang2020g3an}.
However, existing approaches, while separately learning motion and appearance to some extent, fail to learn the true dynamics of motion as they all treat videos as a sequence of frames bound by discretized, fixed-interval timesteps.
Such rigid treatment of the temporal aspect limits the model's capacity to learn natural motions as indicated in Fig.~\ref{fig:motivation}, which eventually leads to generating videos of suboptimal quality.

In this paper, we propose Motion Ordinary Differential Equation GAN (\model), a novel two-stage UVG model that separately learns motion and appearance distributions. In the first stage, the motion generator is responsible for converting the motion noise vector to a sequence of keypoints (\emph{i.e.}, representation of motion). 
To learn continuous-time dynamics of a physical body, we employ a neural ODE~\cite{chen2018neural}, which is a continuous-time model by interpreting the forward pass of the neural networks as solving an ODE.
% We employ a neural ODE for this part, a framework best suitable for learning continuous-time dynamics of a physical body. 
With neural ODE, our motion generator can produce a motion in an arbitrary frame rate, which is especially useful for generating non-linear motion dynamics (\eg, sports) where higher frame rates at certain segments can help the viewer understanding.
In the second stage, given a sequence of keypoints and an appearance noise vector, the motion-conditioned video generator synthesizes a video sequence by combining the two.
With this two-stage approach, \model provides full control of the spatio-temporal aspects of the generated video, such as mixing the motion and appearance from two different datasets. This lends more power to the user to generate diverse videos that even may not exist in the real world.
% (\eg, child running like an adult).

To sum up, \model not only generates high-quality videos with continuous motion dynamics, but also learns completely independent motion density function and appearance density function. We summarize our contributions to the domain of UVG as follows:
\vspace{-0.1cm}
\begin{itemize}
    \setlength{\itemsep}{-3pt}
    \item We present a \emph{novel two-stage unconditional video generation framework} \model, which demonstrates high-quality videos in terms of pixel distribution as well as motion smoothness.
    \item By employing the neural ODE to generate continuous-time motions, \model is able to \emph{dynamically change the frame rate} even when generating a single video sample.
    % \item \model learns completely independent density functions for motion and appearance, enabling \emph{full control of the spatio-temporal aspects} of the generated video.
    \item \model learns completely independent density functions for motion and appearance, enabling \emph{disentanglement of the spatio-temporal aspects} of the generated video.
\end{itemize}
% \vspace{-0.5cm}
\vspace{-0.1cm}
\section{Related Work}
\vspace{-0.1cm}

\textbf{Video Generation from Random Noise.}
\enskip
The goal of unconditional video generation is to learn a mapping function that generates a realistic video given a random noise vector.
Existing approaches tried to decompose a video into several independent components.
In an earlier study, VGAN~\cite{vondrick2016generating} decomposed a video into a foreground object and a background during video synthesis.
In addition, TGAN~\cite{saito2017temporal} tried to split each frame into a fast and slow part.
% and TGAN-v2~\cite{saitotrain} was proposed to further enhance video quality via efficient training while taking a similar approach to decompose.
MoCoGAN~\cite{tulyakov2018mocogan} was the first approach to divide the video signal into appearance and motion.
Lastly, G$^3$AN~\cite{wang2020g3an} proposed a three-stream video generator to promote the disentanglement of appearance and motion to improve video quality.
Our work is distinguished from previous approaches by learning completely independent density functions for motion and appearance where the former is modeled in continuous-time via neural ODE.
This choice or architecture enables not only improved video quality with smooth motion dynamics in arbitrary frame rates, but also transferring motion from one dataset to another.

\noindent \textbf{Conditional Video Generation with Additional Input.}
\enskip
Generating videos with the additional inputs such as semantic segmentation~\cite{wang2018video,wang2019few,pan2019video}, pose  keypoints~\cite{chan2019everybody, siarohin2019first,siarohin2019animating} relates to learning the marginal distributions instead of modeling the joint distributions~\cite{pu2018jointgan}.
There exist a large body of work, where given a single video frame~\cite{yang2018pose,li2018flow,kim2019unsupervised} or a sequence of frames~\cite{wang2018eidetic,wang2019memory,kwon2019predicting}, the model predicts the in-between frames or future frames.
Some works among them are related to our work in that they predict video frames by extracting the pose keypoints from an input image~\cite{yang2018pose, kim2019unsupervised}.
Although these works for conditional video generation also handle video data, our contribution lies in learning separate density functions for motion and appearance, rather than making predictions given an initial video frame(s).

\noindent \textbf{Neural ODE.}
\enskip
Neural ODEs~\cite{chen2018neural} represent one of the continuous-depth deep learning models which employ a neural network to model the dynamics (\textit{i.e.}, vector field) of the latent state. 
Equipped with widely used numerical solvers such as Runge-Kutta and Dormand–Prince method, neural ODE has the capacity to express the latent state in continuous-depth, or equivalently continuous-time.
The continuous nature of neural ODE paved a way to design the continuous time-series modeling as shown in following studies~\cite{rubanova2019latent,yildiz2019ode2vae,de2019gru,dupont2019augmented,park2020vidode}.
Latent ODE~\cite{rubanova2019latent} introduced ODE-RNN as an encoder and demonstrated the effectiveness of handling the time-series data taken at non-uniform intervals.
Furthermore, ODE$^2$VAE~\cite{yildiz2019ode2vae} and Vid-ODE~\cite{park2020vidode} performed continuous-time video prediction conditioned on input video frames, demonstrating the potential to apply neural ODE to computer vision.

\vspace{-0.1cm}
\section{Method}
\label{sec:method}
\vspace{-0.1cm}

\textbf{Problem Statement.}
\enskip
Our task is unconditionally generating a video $\hat{\bv}_{1:T} \in \bbR^{T \times 3 \times H \times W}$ given two noise vectors, the motion noise vector $\bz_m \in \calZ_{M}$ and the appearance noise vector $\bz_a \in \calZ_{A}$, where $T$ denotes the number of frames, $H$ and $W$ the height and width of the generated image, respectively.

\noindent \textbf{Model Overview.} 
\enskip
%Given two noise vectors, the motion noise vector $\bz_m \in \calZ_{M}$ and the appearance noise vector $\bz_a \in \calZ_{A}$, our goal is to generate a video $\hat{\bv}_{1:T} \in \bbR^{T \times 3 \times H \times W}$, where ($T, H, W$) denote the number of frames, height and width respectively.
%To achieve this goal, as introduced in Section~\ref{sec:intro},
We employ a two-stage approach with two separate components: the \emph{motion generator} and the \emph{motion-conditioned video generator}.
% We define the motion as a sequence of keypoints coordinates, which represents the movement of an object.
Starting from $\bz_m$, the motion generator creates a sequence of keypoints, which conveys a plausible movement of an object.
Given the sequence of keypoints and $\bz_a$, the motion-conditioned video generator synthesizes a realistic video following the geometric information in the keypoints.
%Accordingly, the training process consists of two stages, training the motion generator and training the motion-conditioned video generator.
% In the following, we elaborate on the background knowledge about ODE formulation and describe each stage in detail.
In the following, we describe each stage in detail.

\subsection{Stage \Romannum{1}: Motion Generation}
\label{label:stage1}

\begin{figure*}[t!]
    \centering
    \includegraphics[width=0.9\linewidth]{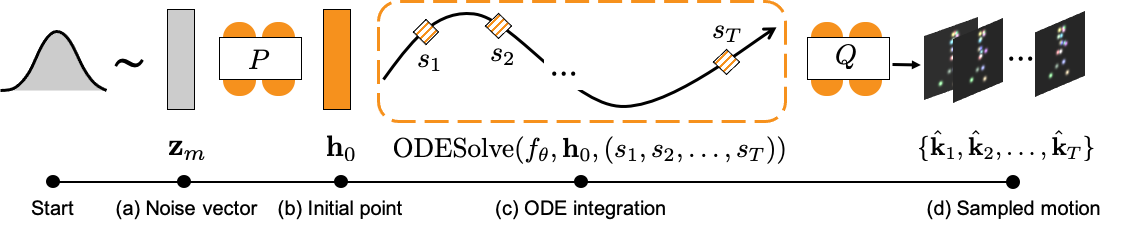}
    \vspace{-0.5cm}
    \caption{\textbf{Overview of Stage \Romannum{1}}
    \enskip
    Motion generator creates a sequence of keypoints representing the motion of an object. 
    (a) Sample a noise vector $\bz_m$ from $\calN(\mathbf{0}, I)$.
    (b) $\bz_m$ is mapped to an initial point $\bh_{0}$ via function $P$.
    (c) By integrating $\bh_{0}$ over the target timesteps $\{s_{1},s_{2},\ldots,s_{T}\}$, latent states for each timestep $\bh_{1:T}$ are obtained.
    (d) Latent states are projected via function $Q$ to a set of evolving keypoints $\hat{\bk}_{1:T}$.}
    \label{fig:stage1_overview}
    \vspace{-0.5cm}
\end{figure*}

%\textbf{Overview.} \enskip
Fig.~\ref{fig:stage1_overview} depicts the overall process of motion generation via neural ODE. 
The motion generator aims to learn a distribution of sequential 2D keypoints coordinates $\bk_{1:T} \in \bbR^{T \times K \times 2}$, where $K$ denotes the number of keypoints, beginning with a noise vector $\bz_m$ drawn from $\calN(\mathbf{0}, \mathbf{I})$.
As an initial step, $\bz_m$ is fed into the initial value mapper $P$ to generate the initial value $\bh_{0}$ for the ODE solver.
Integrating over the target timesteps $\{s_{1},s_{2},\ldots,s_{T}\}$, the ODE solver produces a sequence of latent states $\bh_{1:T}$ which are then transformed into the a sequence of keypoints $\hat{\bk}_{1:T} \in \bbR^{T \times K \times 2}$ via a fully connected neural network $Q$. 
% \jsedit{It is not clear what $P$ is. Can you elaborate more? (Also, it will be nicer to mark $P$ somewhere in Figure 2.)}
Overall, the motion generator is described as
\begin{align}
    \bh_{0} &= P(\bz_{m}), \nonumber\\
    \bh_{1}, \bh_{2}, \ldots, \bh_{T}  &= \text{ODESolve}(f_{\theta}, \bh_{0}, (s_{1}, s_{2}, \ldots, s_T)), \nonumber \\
    \text{each} \quad \hat{\bk}_{t} &= Q(\bh_{t}) \quad t=1,2,\ldots,T,
\end{align}
where $f_\theta$ indicates a fully-connected neural network to approximate $d\bh_{t}/dt$. 
%and $\calS \equiv \{s_{1},s_{2},\ldots,s_{T}\}$ represents a set of target timesteps to solve.
A straightforward way to train the motion generator is to learn the distribution of real coordinates $\bk_{1:T}$ via adversarial training.
However, we observed that the 2D coordinates of the keypoints led to unstable training, and therefore used 2D Gaussian heatmaps as an alternative representation.
These $K$ Gaussian heatmaps $\calH_{t} \in \mathbb{R}^{K \times H \times W}$ are obtained by utilizing a Gaussian-like function centered at $\bk_{t}$, which can be formulated as
% \jsedit{please specify dimensionality of $\calH^{(c, \bu)}_{t}$ here}
\begin{align}
    \calH^{(c, \bu)}_{t} = \exp \Big( -\frac{1}{2\sigma^{2}} \| \bu - \bk^{c}_{t} \|^{2} \Big),
\end{align}
where $\bu \in \Omega$ is the pixel coordinates and $\bk^{c}_{t}$ indicates the coordinates of the $c$-th keypoint at time $t$.

\begin{figure*}[t!]
    \centering
    \includegraphics[width=0.9\linewidth]{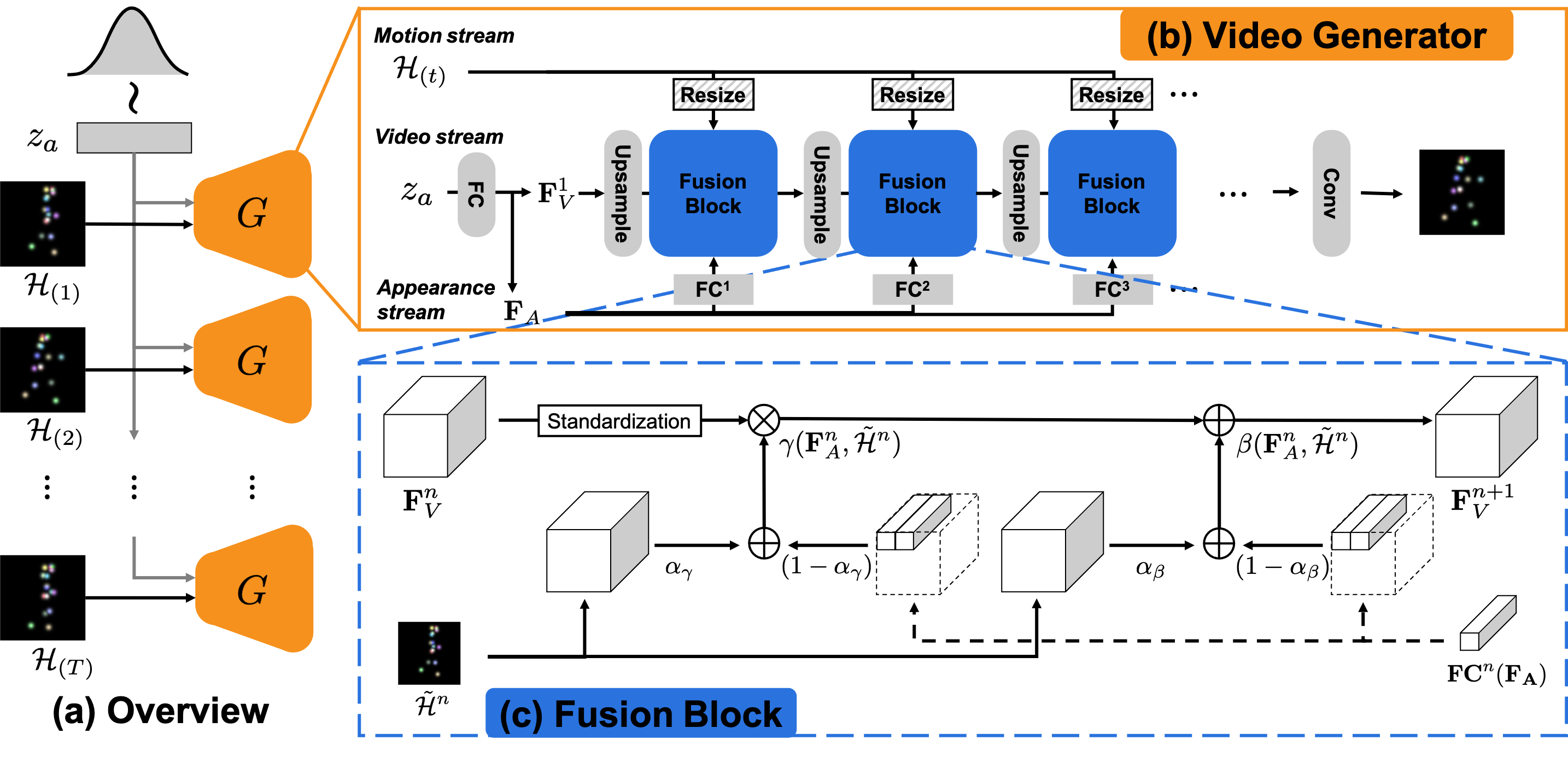}
    \caption{
    \textbf{Overview of Stage \Romannumm{2}}
    \enskip
    % (a) Given an appearance noise vector $\bz_{a}$ and a sequence of Gaussian keypoint heatmaps $\calH_{t}$ from Stage \Romannum{1}, the motion-conditioned video generator $G$ synthesizes a realistic video.
    % (b) The video generator $G$ consists of three streams: motion (top), video (middle), and appearance (bottom).
    % At first, $\bz_a$ is fed into a neural network to produce an appearance feature vector $\mathbf{f}_{a}$.
    % Then $\mathbf{f}_{a}$ is reshaped to an initial video feature map $\bM^{1}_{t}$. Afterwards, a series of $N$ up-sampling layers and composition blocks are used to generate a single video frame at timestep $t$.
    % In particular, the composition block fuses the motion and appearance information from each stream.
    % (c) In the composition block, the resized Gaussian heatmap $\tilde{\calH}^{n}_{t}$ and a projected appearance feature $\mathbf{FC}^{n}(\mathbf{f}_{a})$ are used together to modulate the video feature  $\bM^{n}_{t}$ by estimating an affine transformation.
    % Concretly, $\tilde{\calH}^{n}_{t}$, $\mathbf{FC}^{n}(\mathbf{f}_{a})$ are differently transformed and combined via learnable parameters $\alpha_{\gamma}$ and $\alpha_{\beta}$.
    % Through these parameters, the networks can determine how much of the appearance to apply for this block.
    % Finally, obtained affine parameters $\gamma, \beta$ are applied to $\bM^{n}_{t}$, transforming it according to the motion and appearance information.}
    (a) Given an appearance noise vector $\bz_{a}$ and a sequence of Gaussian keypoint heatmaps $\calH_{t}$ from Stage \Romannum{1}, the motion-conditioned video generator $G$ synthesizes a realistic video.
    (b) The video generator $G$ consists of three streams: motion (top), video (middle), and appearance (bottom).
    At first, $\bz_a$ is fed into a neural network to produce an appearance feature vector, which is reshaped to an initial video feature map $\bM^{1}_{t}$.
    Afterwards, a series of $N$ up-sampling layers and composition blocks are used to generate a single video frame at timestep $t$.
    (c) In the composition block, the resized Gaussian heatmap $\tilde{\calH}^{n}_{t}$ and a projected appearance feature $\mathbf{FC}^{n}(\mathbf{f}_{a})$ are used together to determine learnable parameters $\alpha_{\gamma}$ and $\alpha_{\beta}$, which are applied to 
    the video feature $\bM^{n}_{t}$, transforming it according to the motion and appearance information.
    }
    \label{fig:stage2_overview}
    \vspace{-0.6cm}
\end{figure*}

\noindent \textbf{Loss Function.}
\enskip
We use an adversarial loss to minimize the discrepancy between the distribution of the real keypoints sequences and that of the generated ones. 
For this purpose, we employ two discriminators $D^{\text{(\Romannum{1})}}_\text{fr}, D^{\text{(\Romannum{1})}}_\text{sq}$, where each receives Gaussian heatmaps in individual frame and sequence level, respectively.\footnote{Detailed descriptions about adversarial losses are provided in the supplementary material.}
% \begin{align}
%     &\calL^{\text{(\Romannum{1})}}_\text{adv} = 
%     \bbE_{\calH_{t},\hat{\calH}_{t}} \big[\log D^{\text{(\Romannum{1})}}_\text{fr}(\calH_{t}) + \log (1-D^{\text{(\Romannum{1})}}_\text{fr}(\hat{\calH}_{t})) \big] \\
%     &+ \bbE_{\calH_{1:T}, \hat{\calH}_{1:T}} \big[\log D^{\text{(\Romannum{1})}}_\text{sq}(\calH_{1:T}) + \log (1-D^{\text{(\Romannum{1})}}_\text{sq}(\hat{\calH}_{1:T}))\big]. \nonumber
% \end{align}
% In particular, we employ the WGAN-GP~\cite{gulrajani2017improved} loss as the adversarial loss.
Since ODE integration is solely determined by the initial value, the initial values $\bh_{0}$ need to be diverse in order to generate diverse motions.
To this end, we add an diversity loss of initial value $\calL^{\text{initial}}_\text{div}$ to enforce two different motion noise vectors $\bz_{m}, \bz'_{m} \in \calZ_{M}$ to be embedded in two different places in the initial value space.
\begin{align}
    \calL^{\text{initial}}_\text{div} = \frac{\|\bz_{m} - \bz'_{m}\|_{1}}{\|P(\bz_{m}) - P(\bz'_{m}) \|_{1}}
\end{align}
The overall objective function for stage \Romannum{1} is formulated as
\begin{align}
\label{eq:stage_1_loss}
    \min_{P, Q, f_{\theta}} \max_{D^{\text{(\Romannum{1})}}_\text{fr}, D^{\text{(\Romannum{1})}}_\text{sq}} \calL^{\text{(\Romannum{1})}}_\text{adv} + \lambda^{\text{initial}}_\text{div} \calL^{\text{initial}}_\text{div},
\end{align}
where $\calL^{\text{(\Romannum{1})}}_\text{adv}$ denotes the adversarial loss for stage $\text{\Romannum{1}}$ and $\lambda^{\text{initial}}_\text{div}$ is a hyperparamter controlling the relative importance between the two losses.

\subsection{Stage \Romannumm{2}: Motion Conditioned Video Generation}
%\textbf{Overview} \enskip
As shown in Fig.~\ref{fig:stage2_overview}(a), given a series of keypoints represented as Gaussian heatmaps $\calH_{1:T}$, the motion-conditioned video generator $G$ synthesizes a video $\hat{\bv}_{1:T}$ where the object follows the geometric information of the keypoints sequence. 
Inspired by G$^3$AN~\cite{wang2020g3an}, we use three parallel streams: a motion stream, a video stream, and an appearance stream.
The three streams undergo $N$ number of upsample-then-compose blocks to generate the final video, as depicted in Fig.~\ref{fig:stage2_overview}(b).
Specifically, the appearance noise vector $\bz_{a}$ sampled from $\calN(\mathbf{0}, I)$ is fed into a fully-connected neural network to produce an appearance feature vector $\mathbf{f}_{a} \in \bbR^{L}$, used mainly in the appearance stream to inject appearance information to the video stream.
In the video stream, the initial video feature map $\bM^{1}_{1:T} \in \bbR^{\frac{L}{4} \times T \times 2 \times 2}$ is obtained by reshaping $\mathbf{f}_{a}$, then stacking it across the time axis.
In our implementation, we sequentially upsample the feature map from the initial resolution $2 \times 2$ to the final resolution.
%from  the dimensions of $L$ to $\frac{L}{4} \times 2 \times 2$ and stacking it in the temporal axis.
%\sooedit{Question:C1 이 무엇인가요? ( Composition Block 에서 설명한 것 처럼 channel dimension 이라면 여기서도 한번 언급해 주는 것이 좋을 듯 합니다.)  %2. 특별히 2x2 로 reshape 해야 할 이유가 있을까요? }
%\kyedit{1) $C_1$에 대한 설명을 보충했습니다. 2) 좀 더 일반적으로 쓰면 $H / 2^{N}=2, N=5, H=64$,로 쓸 수도 있을꺼 같습니다. 사실 디자인 초이스이긴안데, 현재 모델에서는 $2 \times 2$에서 출발해서 upsample(x2) 블락을 5번 거쳐서 최종적으로 64 사이즈의 비디오를 만들고 있습니다.}
%Then the feature map is sequentially upsampled while being modified by the appearance and the motion information from the other two streams.
The motion stream injects the motion information from the Gaussian heatmaps into the video stream where at each injection, the $H \times W$ resolution heatmaps are resized to the spatial size of $\bM^{n}_{1:T}$, enabling the model to consider the motion information in multi-scale.
The injection of appearance and motion information is conducted inside the composition block.
%During the forward process, the video feature map $\bF^n_V$ evolves in spatially hierarchical manner for $n > 1$, while $\bF_A$ and a resized Gaussian heatmap $\tilde{\calH}^n$ of the target output width and height are used to transform $\bF^n_V$.
% \jsedit{what resize? what dim? why? more details may be needed.}), combines 

\noindent \textbf{Composition Block.}
\enskip
The composition block takes three inputs $\mathbf{f}_a$ (appearance feature vector), $\bM^{n}_{1:T}$ (video feature map),
%\jsedit{It is unclear why we need both $\bF_A$ and $\bF^{n}_{V,1:T}$ at the same time, given that the latter is simply a reshaped tensor of the former.}
%\kyedit{첫번째 Composition block 같은 경우는 $\bF^{n}_{V,1:T}$ 가 필요로 하지 않는데, 뒤쪽 block들에서는 $F_A$ reshape된거랑 $\bF^{n}_{V,1:T}$가 달라서 일관되도록 서술하기 위해서 이렇게 했습니다. }
$\tilde{\calH}^n$ (resized Gaussian heatmaps), and combines them via spatially adaptive denormalization~\cite{park2019semantic} by estimating scale $\gamma$ and shift parameters $\beta$, and produces $\bM^{n+1}_{1:T}$.
It consists of standardization over all frames (\textit{i.e.,} batch size $\times$ $T$ dimension) followed by an adaptive scaling and shift.
Let $(B, C_n, H_n, W_n)$ be the number of frames, the channel dimension, height and the width of $\bM^{n}_{1:T}$. Then, the transformed activation value at each site ($b \in B, c \in C_n, i \in H_n, j \in W_n$) is given by
\begin{align}
    \gamma_{c,i,j}(\mathbf{f}_{a}, \tilde{\calH}^n) \frac{o_{b,c,i,j} - \mu_{c}}{\sigma_{c}} + \beta_{c,i,j}(\mathbf{f}_{a}, \tilde{\calH}^n),
\end{align}
where $o_{b,c,i,j}$ is an activation value at the site before transformation, the scale and shift parameters $\gamma_{c,i,j}, \beta_{c,i,j}$ are weighted sums of motion and appearance information obtained from $\mathbf{FC}^{n}(\mathbf{f}_a)$ and $\tilde{\calH}_{t}^{n}$, and $\mu_c, \sigma_c$ are the mean and standard deviation of $\bM^{n}_{1:T}$ along channel $c$, respectively (See Fig.~\ref{fig:stage2_overview}(c) for details).

\noindent \textbf{Loss Function.} 
\enskip
The motion-conditioned video generator aims at generating a realistic video retaining the motion information from $\calH_{1:T}$.\footnote{Detailed descriptions about adversarial losses are provided in the supplementary material.}
% To this end, we use adversarial losses to achieve both goals: the first discriminator loss $D_{img}$ for generating realistic frames, the second discriminator loss $D_{vid}$ for retaining motion information, and we use the WGAN-GP~\cite{gulrajani2017improved} loss.
% To this end, we use two motion conditional adversarial losses: $D^{\text{(\Romannum{2})}}_\text{fr}$, $D^{\text{(\Romannum{2})}}_\text{sq}$ encourages the model to generate realistic frames and retaining motion information. \begin{align}
%     \calL^{\text{(\Romannum{2})}}_\text{adv} &= 
%     \bbE_{\bv_{t}, \hat{\bv}_{t}, \calH_{t}}\big[\log D^{\text{(\Romannum{2})}}_\text{fr}([\bv_{t};\calH_{t}]) + \log (1 - D^{\text{(\Romannum{2})}}_\text{fr}([\hat{\bv}_{t};\calH_{t}])) \big] \\ 
%     &+ \bbE_{\bv_{1:T}, \hat{\bv}_{1:T}, \calH_{1:T}}\big[\log D^{\text{(\Romannum{2})}}_\text{sq}([\bv_{1:T};\calH_{1:T}]) + \log (1 - D^{\text{(\Romannum{2})}}_\text{sq}([\hat{\bv}_{1:T};\calH_{1:T}])) \big], \nonumber
% \end{align}
% where $[\boldsymbol{\cdot} \hspace{0.05cm} ; \hspace{0.05cm} \boldsymbol{\cdot}]$ denotes concatenation. We adopt WGAN-GP~\cite{gulrajani2017improved} loss similar to stage I.
%make the generated outputs indistinguishable from real videos as well as to reflect the motion information adding Gaussian heatmaps as additional condition to consider.
%Specifically, we adopt two discriminators $D^{img}_{V}, D^{vid}_{V}$ and leverage the WGAN-GP~\cite{gulrajani2017improved} loss.
In addition, we employ the pixel-level diversity loss to make output videos distinctive given two different appearance noise vectors $\bz_{a}, \bz'_{a} \in \calZ_{A}$.
\begin{align}
    \calL^{\text{pixel}}_{\text{div}} = \frac{\|\bz_{a} - \bz'_{a}\|_{1}}{\|G(\bz_{a}; \calH_{1:T}) - G(\bz'_{a}; \calH_{1:T}) \|_{1}}
\end{align}

\noindent The complete objective function for stage \Romannumm{2} is described as
% \begin{align}
%     \calL^{\calV}_{D} &= \calL^{adv}_{D^{img}_{V}} + \calL^{adv}_{D^{vid}_{V}} \nonumber\\
%     \calL^{\calV}_{G} &= \calL^{adv}_{G^{img}_{V}} + \calL^{adv}_{G^{vid}_{V}} + \lambda^{pixel}_{div}\calL^{pixel}_{div}
% \end{align}
\begin{align}
\label{eq:stage_2_loss}
    \min_{G} \max_{D_\text{img}, D_\text{vid}} \calL^{\text{(\Romannumm{2})}}_\text{adv} + \lambda^\text{pixel}_\text{div} \calL^\text{pixel}_\text{div},
\end{align}
where $\calL^{\text{(\Romannumm{2})}}_\text{adv}$ denotes the adversarial loss for stage $\text{\Romannumm{2}}$ and $\lambda^{\text{pixel}}_{\text{div}}$ adjusts the relative importance between losses. 
\vspace{-0.1cm}
\section{Experiments}
\label{sec:exp}
\vspace{-0.1cm}

% We first describe the experiment setup, followed by the quantitative evaluation of \model comparing against baseline methods for unconditional video generation in both fixed frame rate and dynamic frame rate.
% In qualitative analysis, we first study the effectiveness of neural ODE in learning motion dynamics.
% Then, we demonstrate various video generation manipulations to test the versatility and robustness of \model such as fixing motion while varying appearance, transferring motions from one dataset to another, dynamic control of frame rate, and visualizing diverse motions in continuous-time space.
%%%%%%%%%%%%%%%%%%%%%%%%%%%%%%%%%%%%%%%%%%%%%%%%%%%%%%%%%%%%%%%%%%%%%%%%%%%%%%%%%%%%%%%%%%%%%%%%%%%%%%%%%%%

% \subsection{Experimental Setup}
\textbf{Datasets.}
\enskip
We evaluate our method on four datasets:
1) \textit{Weizmann Action}~\cite{blank2005actions} consists of 90 videos of 9 subjects performing 10 actions (\textit{e.g.,} walk, jumping-jack). 
2) From \textit{KTH Action}~\cite{schuldt2004recognizing}, we select videos of 25 subjects performing three types of actions (boxing, hand waving, and hand clapping) always having single person in the video. 
3) \textit{MUG}~\cite{aifanti2010mug} contains 1,254 video sequences of six facial expressions, such as anger, disgust, and happiness.
4) \textit{UvA-NEMO}~\cite{dibekliouglu2012you} is comprised of 1,240 videos of 400 smiling subjects.
We resize the videos to the $64 \times 64$ resolution for all datasets.

% What are the FPS of all datasets?

For training the motion generator, the ground truth keypoints are obtained by using a pre-trained keypoint detector~\cite{fang2017rmpe,li2018crowdpose,bulat2017far}.
In particular, we use 13 out of 68 face keypoints
\footnote{2, 9, 16, 20, 25, 38, 42, 45, 47, 49, 52, 55, 58th facial landmark locations detected using open face alignment library (\url{https://github.com/1adrianb/face-alignment}). We provide a detailed illustration in supplementary material.}
including eyes and nose to represent the facial expression for \textit{MUG} and \textit{UvA-NEMO}. And we use 17 pose keypoints for \textit{Weizmann Action} and \textit{KTH Action}.

\begin{table}[t!]
    \centering
    \small
    \begin{tabular}{lcccc}
        \toprule
        & Weizmann & MUG & UvA & KTH \\
        \midrule
        VGAN~\cite{vondrick2016generating} & 99.03 & 104.71 & 103.70 & 103.31 \\
        TGAN-v2~\cite{saitotrain} & 83.90 & 72.60 & 86.91 & 65.03 \\
        MoCoGAN~\cite{tulyakov2018mocogan} & 93.93 & 35.12 & 49.58 & 36.90 \\
        G$^3$AN~\cite{wang2020g3an} & 64.07 & 21.76 & 39.24 & 44.84 \\
        \midrule
        \model & \textbf{55.06} & \textbf{17.90} & \textbf{15.38} & \textbf{28.28} \\
        \bottomrule \\
    \end{tabular}
    \vspace{-0.2cm}
    \caption{Video FID scores on 4 datasets. Lower values are better.}
    \label{Table:quantitative comparison}
    \vspace{-0.4cm}
\end{table}

\begin{figure*}[h!]
    \centering
    \includegraphics[width=\linewidth]{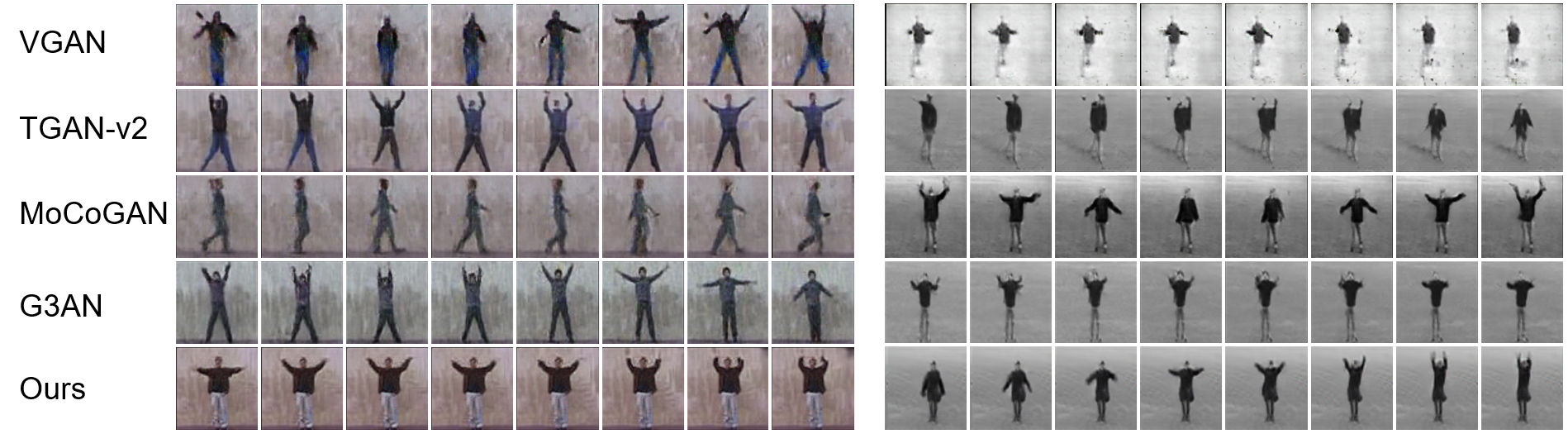}
    \vspace{-0.5cm}
    \caption{Qualitative comparison with baselines on human action datasets.}
    \label{fig:qualitative baseline}
    \vspace{-0.2cm}
\end{figure*}

% \begin{figure}[t!]
%     \centering
%     \includegraphics[width=0.8\linewidth]{figures/figure_continuous_fid.pdf}
%     \vspace{-0.5cm}
%     \caption{FID scores with varied frame rates on Weizmann Action. \model consistently outperforms MoCoGAN, showing marginal performance drop to generate at higher frame rate.}
%     \vspace{-0.5cm}
%     \label{fig:continuous_fid}
% \end{figure}

\noindent \textbf{Evaluation on Fixed Frame Rate.}
\enskip
We compare our model with unconditional video generation baselines~\cite{saitotrain,tulyakov2018mocogan,wang2020g3an,vondrick2016generating} on four datasets. We followed the hyperparameters as presented in the papers.
Table~\ref{Table:quantitative comparison} shows qualitative comparison of \model with VGAN~\cite{vondrick2016generating}, TGAN-v2~\cite{saitotrain}, MoCoGAN~\cite{tulyakov2018mocogan} and G$^{3}$AN~\cite{wang2020g3an} on four different datasets by measuring video Fréchet Inception Distance (FID)~\cite{wang2020g3an}, which is a widely used metric for evaluating the quality of videos.
As seen in Table~\ref{Table:quantitative comparison}, \model consistently outperforms all baseline models.
This indicates our synthetic videos are not just visually realistic but keeps the same realistic quality consistently across time compared to the baselines.
Also, Fig.~\ref{fig:qualitative baseline} shows visual comparisons using human action dataset, where MODE-GAN produces the competitive results compared to the baseline models.\footnote{Other visual comparisons using facial expression dataset are shown in the supplementary material.}
This state-of-the-art performance shows that explicitly decomposing the video generation process into spatial (\textit{i.e.} appearance) and temporal (\textit{i.e.} motion) components, and modeling the latter in the natural continuous-time leads to more realistic synthetic outcomes.

\begin{figure}[t!]
    \centering
    \subfigure[Recurrent neural network]{
        % \label{fig:a}
        \includegraphics[width=0.48\textwidth]{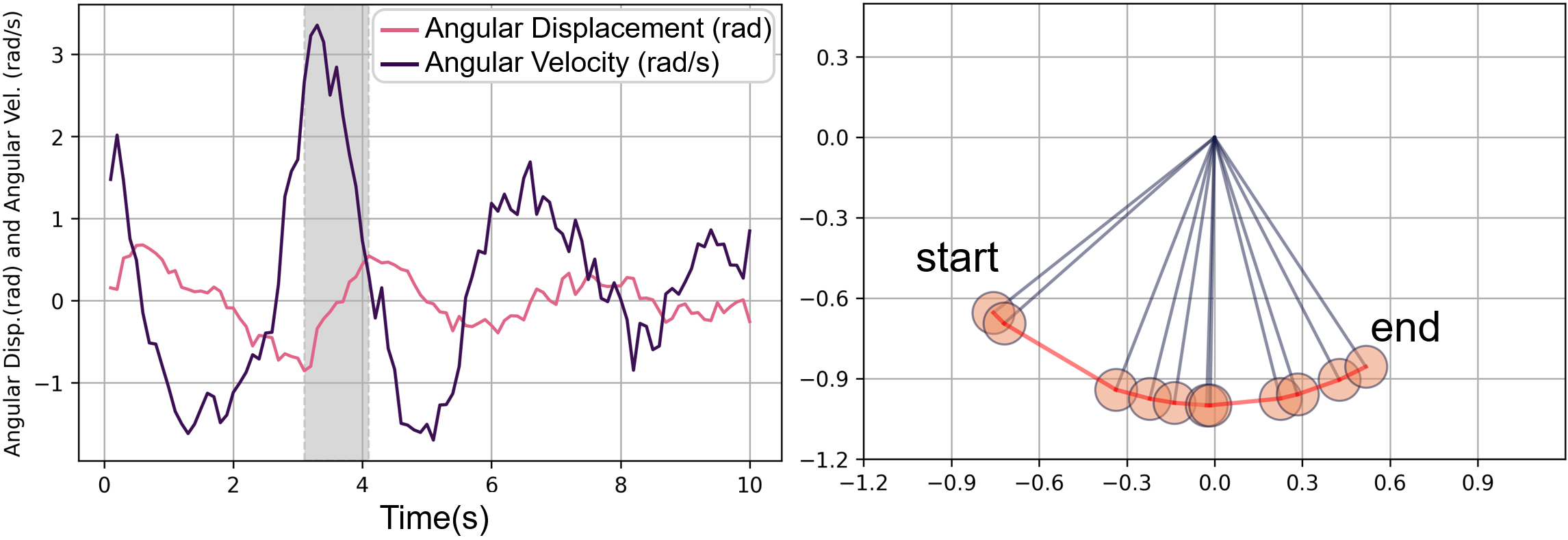}
    }
    \subfigure[Neural ODE]{
        % \label{fig:b}
        \includegraphics[width=0.48\textwidth]{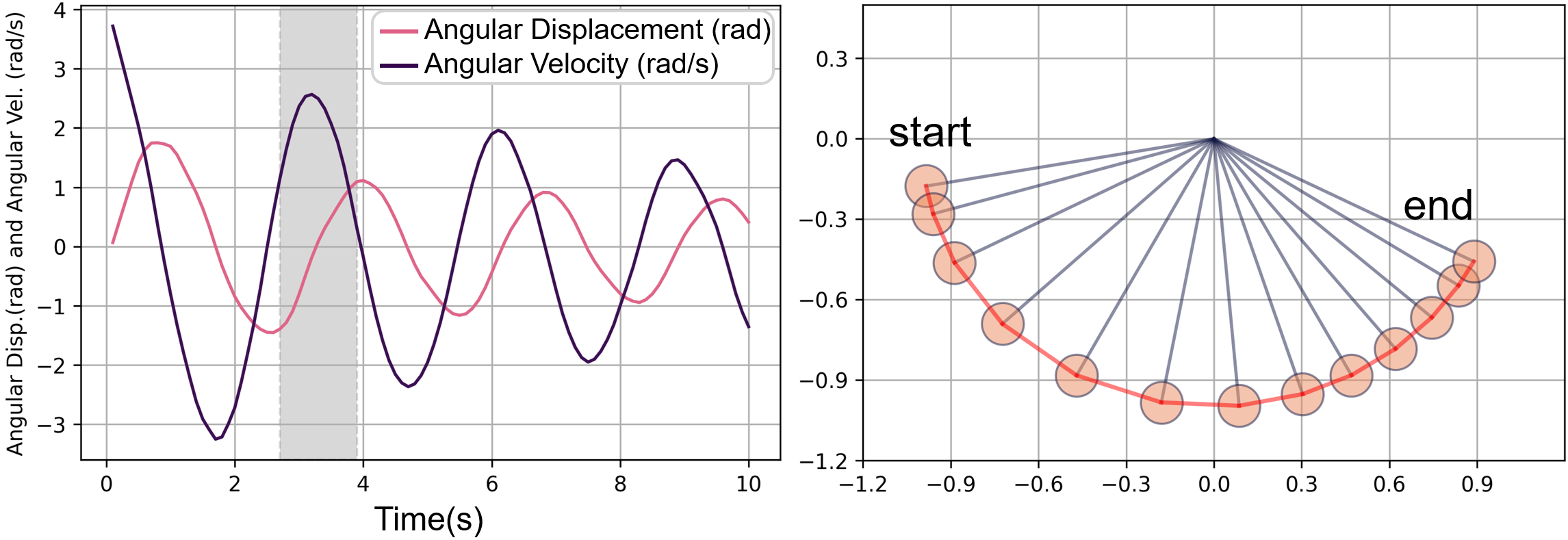}
    }
    \caption{Comparison of generated pendulum dynamics (\textit{Left}: sampled $\theta$ and $\dot{\theta}$, \textit{Right}: visualization of the bob movement in the \textcolor{gray}{gray} area of the left. (a) RNN fails to preserve the smoothness of the pendulum movement ($\theta$ and $\dot{\theta}$ severely fluctuate across time), yielding a rigid and irregular movement of the bob (b) Neural ODE successfully mimics the pendulum movement maintaining continuous and smooth bob motion.}
    \label{fig:pendulum}
    \vspace{-0.5cm}
\end{figure}

\noindent \textbf{Effectiveness of ODE in Learning Dynamics.}
\enskip
We compare the effectiveness of neural ODE with recurrent neural networks (RNNs) in learning the distribution of dynamics starting from a random noise.
Motivated by previous work~\cite{johansson1973visual} that interprets human motion as a relative \textit{pendulum} dynamics of keypoints, we take a \textit{pendulum} system which is mathematically formulated as
\begin{align}
    \ddot{\theta} + \bigg(\frac{B}{M}\bigg) \cdot \dot{\theta} + \bigg(\frac{g}{L}\bigg) \cdot \sin(\theta)=0,
\end{align}
where $\theta$, $B$, $g$, $L$, and $M$ are the angular displacement, damping factor, gravity force, length of pendulum, and mass of bob, respectively.
Our goal is to train the generator which is capable of simulating the dynamics of the pendulum system. 
For this purpose, the model aims at producing $\theta$ and $\dot{\theta}$, a plausible physical parameters for the pendulum system.
In experiment, $g$ is a constant and fixed to 9.81 $\text{m/s}^2$, whereas $B$, $L$, and $M$ are stochastically determined by sampling each factor from Gaussian distribution with means of 0.2, 1.0, 1.0 and unit variance, respectively.
As a baseline, we employ an RNN-based motion generator, where we substitute the neural ODE in stage~\Romannum{1} with an RNN.
The RNN takes Gaussian noises at each time step and aims to generate a plausible $\theta$ and $\dot{\theta}$.

\begin{figure}[t!]
    \centering
    \subfigure[Weizmann Action]{
    \includegraphics[width=0.49\textwidth]{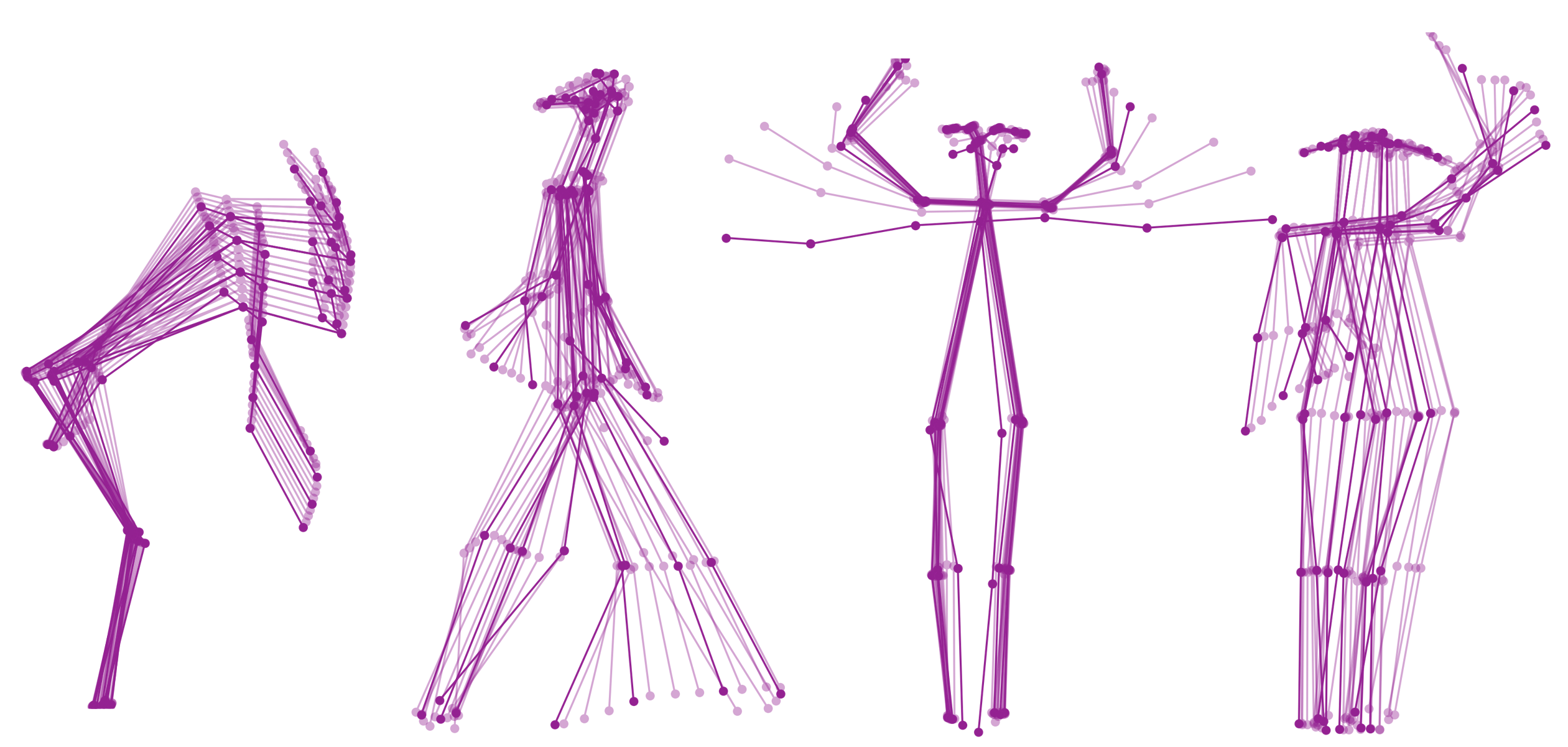}}
    \subfigure[KTH Action]{
    \includegraphics[width=0.49\textwidth]{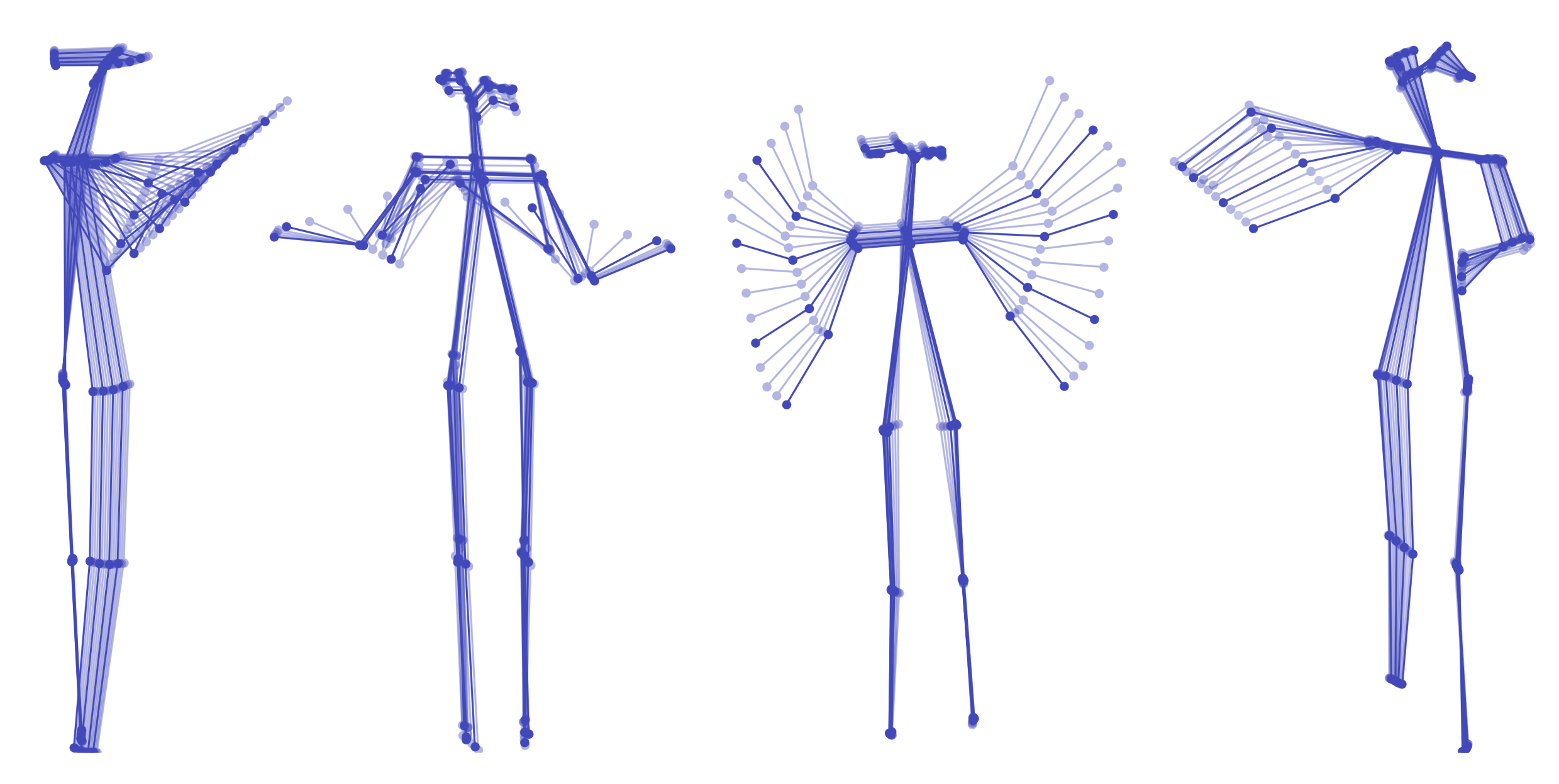}}
    \caption{Examples of sequence of continuously evolving keypoints: We sample a sequence of 64 keypoints from \model trained at 16 FPS. To visualize the keypoints dynamics, every 4th frame (\textit{i.e.} 16 / 64) are marked in bold, while other frames are  faintly illustrated.}
    \label{fig:continuous_keypoints}
    \vspace{-0.5cm}
\end{figure}

As seen in Fig.~\ref{fig:pendulum}(a), the RNN-based motion generator fails to generate smooth dynamics, running off the expected trail of the pendulum dynamics.
On the other hand, as in Fig~\ref{fig:pendulum}(b), our ODE-based motion generator successfully simulates the pendulum dynamics.
This demonstrates that neural ODE has a benefit in learning the continuous motion dynamics compared to RNN. 
As seen in Fig.~\ref{fig:pendulum}, RNN fails to smoothly interpolate the pendulum trajectory (a) whereas neural ODE successfully simulates the smooth dynamics of the pendulum (b), which indicates the potential of neural ODE for learning the dynamics of real-world videos smoothly.
In other words, the capacity of learning the smooth dynamics can be extended to simulate the unseen dynamics between two frames in more natural manner without rather unnatural rigid and irregular movements. 
Such capability can be seen in Fig.~\ref{fig:continuous_keypoints}, where the motion generator successfully fills the plausible keypoints at unseen timesteps.
Detailed descriptions about data generation process and model architectures are provided in supplementary material.

\noindent \textbf{Diverse Motions in Continuous-time Space.}
\enskip
The motion generator is to learn the distribution of motions, thereby generating a plausible motion.
Specifically, neural ODE allows the motion generator to model dynamics of keypoints in continuous-time domain.
%As a result, we can visualize diverse and continuously solved keypoints representations describing motion (even higher than trained FPS).
Fig.~\ref{fig:continuous_keypoints} shows keypoints representations of various motions as we integrate densely across time, trained on \textit{Weizmann Action} and \textit{KTH Action} datasets.
We see the sampled motions are natural and fluid, demonstrating that the motion generator has successfully learned the underlying distribution of video motions.

\begin{figure}[t!]
    \centering
    \subfigure[Fixed Motion]{
    \includegraphics[width=0.49\textwidth]{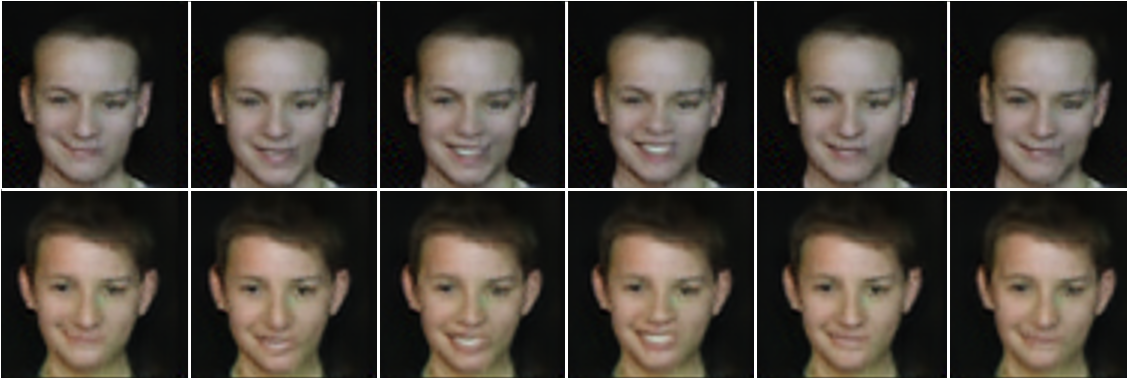}}
    \subfigure[Fixed Appearance]{
    \includegraphics[width=0.49\textwidth]{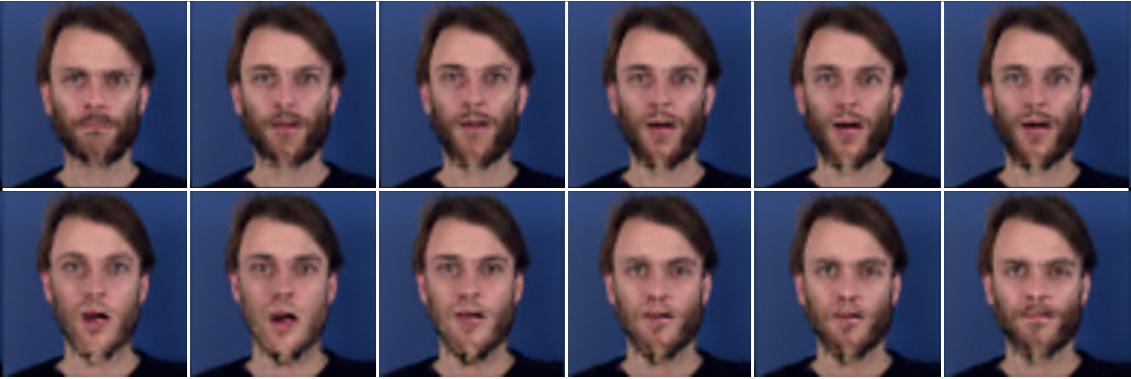}}
    \caption{An example of manipulating motion and appearance: (a) Fixed motion (facial expression) with different appearances, (b) Fixed appearance with different motions.}
    \label{fig:disentangle}
    \vspace{-0.5cm}
\end{figure}

\noindent \textbf{Motion and Appearance Manipulation.}
\enskip
Recall that our model synthesizes a video by combining separate representations for \textit{motion} and \textit{appearance}. In this experiment, we demonstrate our model's capability to generate videos while maintaining one component (\textit{i.e.,} motion \textit{or} appearance), by fixing either $\bz_m$ or $\bz_a$ and varying the other.

As shown in Fig.~\ref{fig:disentangle}, we observe different combinations can successfully produce videos preserving the fixed component, while altering the other.
These results indicate that we can successfully manipulate the video generation process by controlling appearance and motion independently.

\begin{figure}[t!]
    \centering
    \subfigure[MUG: \textit{Surprise} $\Rightarrow$ UvA-NEMO]{
    \includegraphics[width=0.49\textwidth]{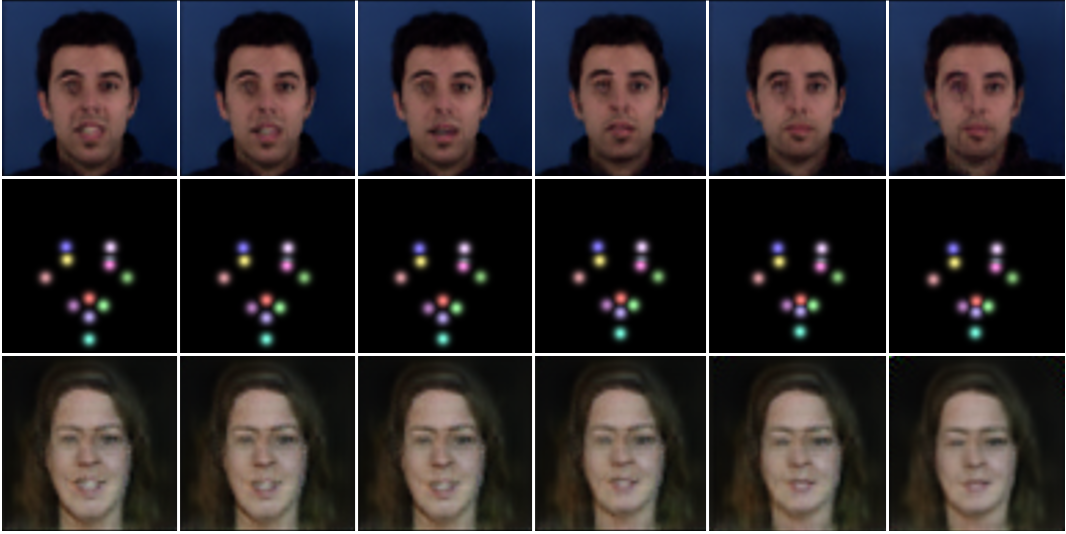}}
    \subfigure[Weizmann: \textit{One hand waving} $\Rightarrow$ KTH]{
    \includegraphics[width=0.49\textwidth]{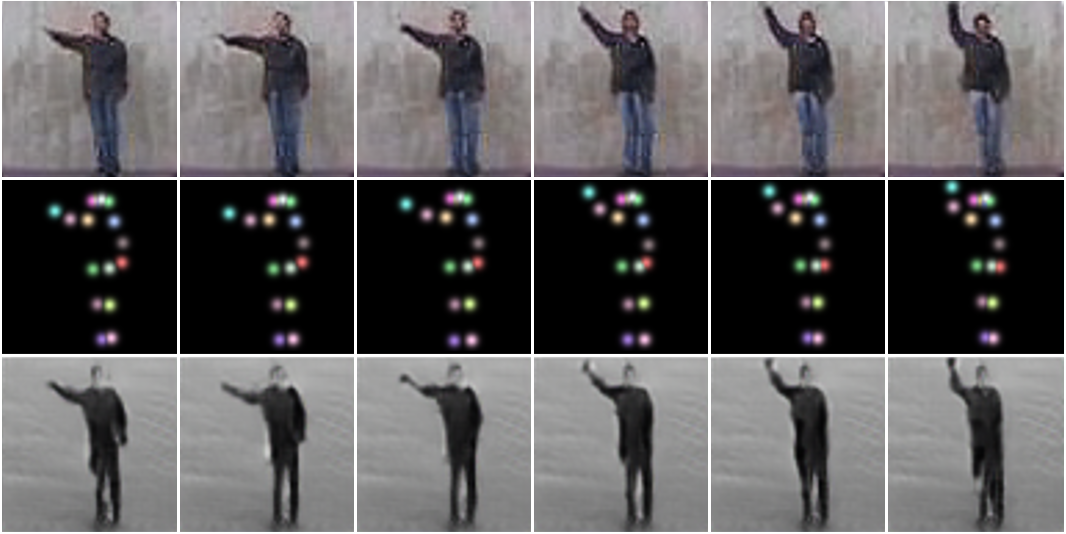}}
    \caption{Motion transfer results across different video domains (\textit{First row}: generated output at source domain, \textit{Second row}: corresponding keypoints sequence, \textit{Third row}: generated output at target domain).}
    \label{fig:motion_trasfer}
    \vspace{-0.4cm}
\end{figure}

% \begin{figure*}[t!]
%     \centering
%     \includegraphics[width=1.0\linewidth]{figures/figure_continuous.png}
%     \caption{Videos generated in diverse frame rates by \model, trained at 16 FPS.
%     %but is able to successfully generate videos in novel frame rates.
%     }
%     \label{exp:continuous_generation}
%     \vspace{-0.5cm}
% \end{figure*}

\noindent \textbf{Motion Transfer between Different Video Domains.} 
One of the distinguishing applications of \model is to import a non-existing motion to a different video domain.
Such application stems from the fact that motion-conditioned video generators share the common motion space expressed as keypoints representations.
Therefore, by connecting the motion generator trained on one video domain with the motion-conditioned video generator trained on another one, we can combine motion and appearance from each video domain. 

Fig.~\ref{fig:motion_trasfer} shows motion transfer examples in facial expression and human action datasets.
(a) Although UvA-NEMO contains only smile motion, \textit{surprise} expression can be imported by adopting the motion model trained on MUG dataset.
(b) In a similar manner, we synthesize the KTH Action domain video using \textit{one hand waving} motion adopted from the Weizmann Action domain.

\begin{figure}[t!]
    \centering
    \subfigure[Weizmann: \textit{Two hand waving}]{
    \includegraphics[width=\textwidth]{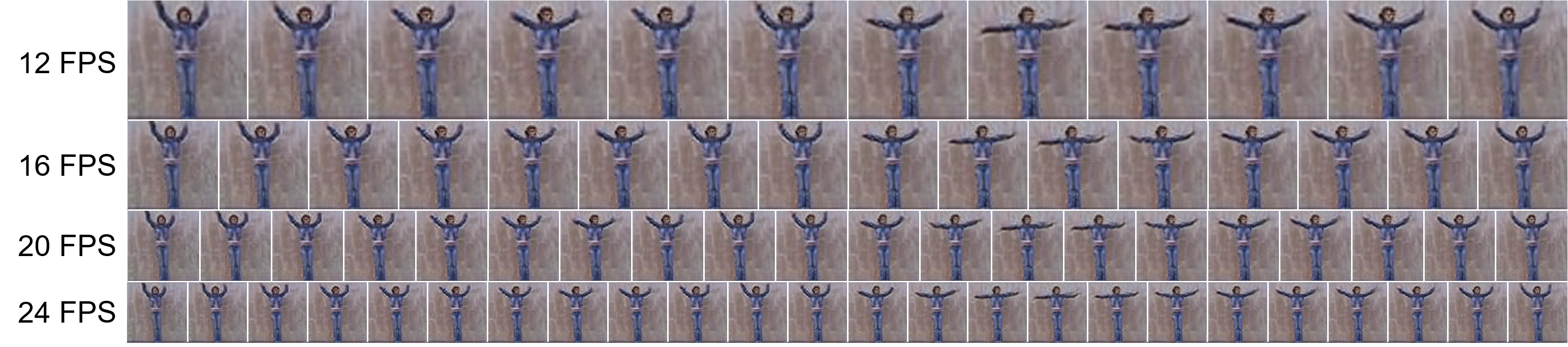}}
    \subfigure[Weizmann: \textit{Bending}]{
    \includegraphics[width=\textwidth]{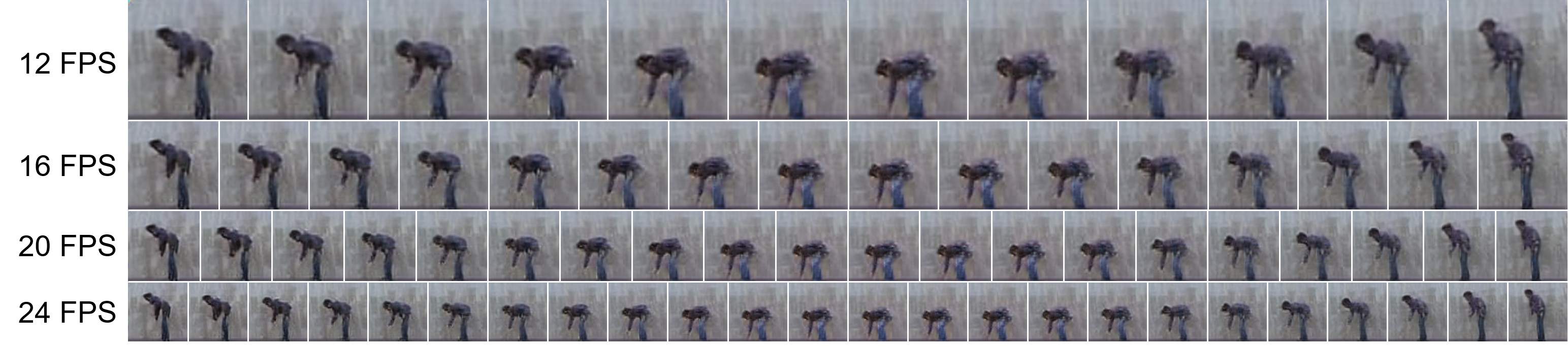}}
    \caption{Videos generated in diverse frame rates by \model, trained at 16 FPS.}
    \label{fig:continuous_generation}
    \vspace{-0.5cm}
\end{figure}

\noindent \textbf{Arbitrary Frame Rate Video Generation.}
\enskip
Another application of our model is to generate videos in arbitrary frame rates based on the continuously generated keypoints from the motion generator.
As illustrated in Fig.~\ref{fig:continuous_generation}, our model is capable of synthesizing a video at various frame rates, densely dividing the continuous-time domain.
Synthesized videos successfully fill in the frame at arbitrary timesteps, demonstrating that our model understand the underlying dynamics of motion (\textit{i.e., two hand waving}).
Furthermore, we can dynamically control the frame rate even for a single video (\textit{e.g.} from slow to fast motion) by controlling integration time span of the motion generator. Such results are provided in supplementary material as a video.

% \ecedit{We should emphasize we can dynamically control the frame rate even for a single video (e.g. slow motion). Do we have examples in the supplementary?}
% \kyedit{Plan to provide in the supplementary material or videos OR we can show it in main manuscript.}

% \textbf{Non-human Object Video Generation}\kyedit{Results are not good enough...}
% \enskip
% Although our model has a strength in continuous and smooth video generation, someone can argue that this might be unsuitable when it comes to handling non-canonical movement. 
% In response, we additionally train our model using Moving GIF (M-GIF) \jsedit{please add citation or URL}, which consists of 1,000 videos including animation animals running or walking
% Fig.~\ref{} shows qualitative results of our model to 

\vspace{-0.1cm}
\section{Summary and Future Work}
\vspace{-0.1cm}

In this paper, we propose a novel framework \model for unconditional video generation.
Based on the observation that real world videos
can be decomposed into spatial (\textit{i.e.} static appearance) and temporal (\textit{i.e.} continuous motion dynamics) aspects, we employ neural ODE to handle the continuous nature of video motion.
In addition, a two-stage approach enables \model to focus on modeling the motion itself, allowing our model to separately learn motion and appearance distributions.
Experimental results not only demonstrate its ability to generate high-quality videos but also its versatile functionality including continuous-time and cross-domain video generation.

% Despite its success, \model still has additional rooms to improve in synthesizing highly complicated or large resolution videos.
Current motion generator of \model focuses on modeling the single person's dynamics, yet it bears a potential to be used for modelling the dynamics of multiple people and complex videos. 
As a future direction, we plan to explore how to generate highly complicated real world semantics in continuous-time domain.

\vspace{-1mm}
\section*{Acknowledgement}
This work was supported by the Institute of Information \& communications Technology Planning \& Evaluation (IITP) grant funded by the Korean government (MSIT) (No. 2019-0-00075, Artificial Intelligence Graduate School Program (KAIST) and No. 2021-0-01778, Development of human image synthesis and discrimination technology below the perceptual threshold) and by Kakao Enterprise.

\bibliography{reference}

%%%%%%%%% For Arxiv
\newpage

\begin{center}
    \vspace*{0.5cm}
    \LARGE{\textcolor{bmv@sectioncolor}{\bf{Supplementary Materials}}}
    \vspace*{0.5cm}
\end{center}
% \noindent\rule{\linewidth}{0.8pt}
\hrule
\vspace*{0.5cm}

\section*{A. Background on ODE Formulation}
The essence of neural ODEs~\cite{chen2018neural} lies in employing a neural network $f$ to estimate the vector field of the latent state $\bh$. 
Formally, $\bh$ in the continuous domain $T \in \calT$ can be represented as
\begin{equation}
  \frac{d\bh(t)}{dt} = f_{\theta}(\bh(t), t), \quad \bh(T) = \bh(0) + \int_{0}^{T} f_{\theta}(\bh(t), t) dt, \nonumber
\end{equation}
where $\theta$ is a set of trainable parameters of $f$. 
Regarding time-series modeling, $\calT$ corresponds to the continuous time domain, thereby allowing the model to characterize the latent state over the continuously evolving time.
In our work, the neural ODE is used to model the continuous-time dynamics of the keypoint sequence, which conveys successive geometric information on the object movement.

% Table for Motion Generator & Discriminator Architecture
\begin{table}[h]
\centering
\parbox{0.45\textwidth}{
    \scriptsize
    \begin{tabular}{@{}cccc@{}}
        \toprule
        \textbf{Part} & \textbf{Layer} & \textbf{Activation} & \textbf{Output Shape}\\
        \midrule \midrule
        $\bz_m$ & - & - & 128 \\
        \midrule
        \multirowcell{4}{Function \\ $P$}
        & Linear & ReLU & 128 \\
        & Linear & ReLU & 128 \\
        & Linear & ReLU & 128 \\
        & Linear & - & 128 \\
        \midrule
        \multirowcell{2}{ODE Solver \\ $f_{\theta}$}
        & Linear & Tanh & 128 \\
        & Linear & - & 128 \\
        \midrule
        \multirowcell{3}{Function \\ $Q$}
        & Linear & ReLU & 128 \\
        & Linear & ReLU & 128 \\
        & Linear & - & 2$\times$K \\
        \midrule
        Heatmaps $\calH_{1:T}$ & - & - & T$\times$K$\times$64$\times$64 \\
        \bottomrule
    \end{tabular}
    \caption{Motion generator architecture of our model.}
    \label{Motion Generator Arch}
}
\hfill
\parbox{0.5\textwidth}{
    \scriptsize
    \begin{tabular}{@{}cccc@{}}
        \toprule
        \textbf{Part} & \textbf{Layer} & \textbf{Activation} & \textbf{Output Shape}\\
        \midrule \midrule
        Heatmap $\calH_t$ & - & - & K$\times$64$\times$64 \\
        \midrule
        \multirowcell{5}{Discriminator \\ $D^{(\Romannum{1})}_{fr}$}
        & Conv & LeakyReLU & 32$\times$32$\times$32 \\
        & Conv & LeakyReLU & 64$\times$16$\times$16 \\
        & Conv & LeakyReLU & 128$\times$8$\times$8 \\
        & Conv & LeakyReLU & 256$\times$4$\times$4 \\
        & Conv & LeakyReLU & 256$\times$1$\times$1 \\
        & Conv & - & 1 \\
        \midrule \midrule
        Heatmaps $\calH_{1:T}$ & - & - & T$\times$K$\times$64$\times$64 \\
        \midrule
        \multirowcell{9}{Discriminator \\ $D^{(\Romannum{1})}_{sq}$}
        & Conv & LeakyReLU & T$\times$32$\times$32$\times$32 \\
        & Conv & LeakyReLU & T/2$\times$32$\times$32$\times$32 \\
        & Conv & LeakyReLU & T/2$\times$64$\times$16$\times$16 \\
        & Conv & LeakyReLU & T/4$\times$64$\times$16$\times$16 \\
        & Conv & LeakyReLU & T/4$\times$128$\times$8$\times$8 \\
        & Conv & LeakyReLU & T/8$\times$128$\times$8$\times$8 \\
        & Conv & LeakyReLU & T/8$\times$256$\times$4$\times$4 \\
        & Conv & LeakyReLU & T/16$\times$256$\times$4$\times$4 \\
        & Conv & LeakyReLU & T/16$\times$256$\times$1$\times$1 \\
        & Conv & - & 1 \\
        \bottomrule
    \end{tabular}
    \caption{Motion discriminator architecture of our model.}
    \label{Motion Discriminator Arch}
}
\vspace{-0.2cm}
\end{table}

\section*{\textsc{B. Implementation details}}\label{1_details}

\subsection*{B.1. Network Architecture}

We provide the architecture details of \model, which consists of two stages.
All architectures are described for the purpose of generating 64$\times$64 videos of 16 timesteps.
The architectures of the motion generator and the discriminator are shown in Tables~\ref{Motion Generator Arch} and \ref{Motion Discriminator Arch}, respectively.
% We adopt the method to create the heatmaps from the keypoint coordinates.
% Gaussian Heatmap을 만드는 방법은 A 논문을 참고하였다.
% 64x64 비디오를 만드는 기준으로 구조가 구성되어있다.
% Discriminator의 LeakyReLU의 slope는 모두 0.2로 설정 / Conv의 kernel size는 4로 설정? stride...
Also, the architectures of the video generator and discriminator are shown in Tables~\ref{Video Generator Arch} and \ref{Video Discriminator Arch}, respectively.

% Table for Video Generator & Discriminator Architecture
\begin{table}[t!]
\centering
\parbox{0.48\textwidth}{
    \scriptsize
    \begin{tabular}{@{}cccc@{}}
        \toprule
        \textbf{Part} & \textbf{Layer} & \textbf{Activation} & \textbf{Output Shape}\\
        \midrule \midrule
        $\bz_a$ & - & - & 128 \\
        \midrule
        \multirowcell{4}{FC}
        & Linear & LeakyReLU & 128 \\
        & Linear & LeakyReLU & 128 \\
        & Linear & LeakyReLU & 128 \\
        & Linear & LeakyReLU & 8192 \\
        \midrule
        \multirowcell{10}{Generator \\ $G$}
        & $\bM^1_t$ & - & 512$\times$4$\times$4 \\
        \cmidrule(lr){2-4}
        & Upsample & - & 512$\times$8$\times$8 \\
        & Comp. Block & - & 512$\times$8$\times$8 \\
        \cmidrule(lr){2-4}
        & Upsample & - & 512$\times$16$\times$16 \\
        & Comp. Block & - & 256$\times$16$\times$16 \\
        \cmidrule(lr){2-4}
        & Upsample & - & 256$\times$32$\times$32 \\
        & Comp. Block & - & 128$\times$32$\times$32 \\
        \cmidrule(lr){2-4}
        & Upsample & - & 128$\times$64$\times$64 \\
        & Comp. Block & - & 64$\times$64$\times$64 \\
        \cmidrule(lr){2-4}
        & Conv3$\times$3 & Tanh & 3$\times$64$\times$64 \\
        \bottomrule
    \end{tabular}
    \caption{Video generator architecture of our model.}
    \label{Video Generator Arch}
}
\hfill
\parbox{0.48\textwidth}{
    \scriptsize
    \begin{tabular}{@{}cccc@{}}
        \toprule
        \textbf{Part} & \textbf{Layer} & \textbf{Activation} & \textbf{Output Shape}\\
        \midrule \midrule
        Frame $\bv_t$ & - & - & 3$\times$64$\times$64 \\
        \midrule
        \multirowcell{5}{Discriminator \\ $D^{(\Romannum{2})}_{fr}$}
        & Conv & LeakyReLU & 64$\times$32$\times$32 \\
        & Conv & LeakyReLU & 128$\times$16$\times$16 \\
        & Conv & LeakyReLU & 256$\times$8$\times$8 \\
        & Conv & LeakyReLU & 512$\times$4$\times$4 \\
        & Conv & LeakyReLU & 512$\times$1$\times$1 \\
        & Conv & - & 1 \\
        \midrule \midrule
        Video $\bv_{1:T}$ & - & - & T$\times$3$\times$64$\times$64 \\
        \midrule
        \multirowcell{9}{Discriminator \\ $D^{(\Romannum{2})}_{sq}$}
        & Conv & LeakyReLU & T$\times$64$\times$32$\times$32 \\
        & Conv & LeakyReLU & T/2$\times$64$\times$32$\times$32 \\
        & Conv & LeakyReLU & T/2$\times$128$\times$16$\times$16 \\
        & Conv & LeakyReLU & T/4$\times$128$\times$16$\times$16 \\
        & Conv & LeakyReLU & T/4$\times$256$\times$8$\times$8 \\
        & Conv & LeakyReLU & T/8$\times$256$\times$8$\times$8 \\
        & Conv & LeakyReLU & T/8$\times$512$\times$4$\times$4 \\
        & Conv & LeakyReLU & T/16$\times$512$\times$4$\times$4 \\
        & Conv & LeakyReLU & T/16$\times$512$\times$1$\times$1 \\
        & Conv & - & 1 \\
        \bottomrule
    \end{tabular}
    \caption{Video discriminator architecture of our model.}
    \label{Video Discriminator Arch}
}
\vspace{-0.5cm}
\end{table}

\subsection*{B.2. Evaluation Metric}
We evaluate all video generation models via Frechet Inception Distance (FID)~\cite{heusel2017gans,wang2018video}, measuring the distance between two sets of videos based on their embeddings from a pre-trained feature extractor~\cite{hara3dcnns}.
Specifically, the video FID~\cite{wang2020g3an} is computed as
\begin{equation}
    \|\mu - \Tilde{\mu}\|^2 + \mathbf{Tr}(\Sigma + \Tilde{\Sigma} - 2 \sqrt{\Sigma \Tilde{\Sigma}}),
\end{equation}
where $\mu$, $\Tilde{\mu}$, $\Sigma$, and $\Tilde{\Sigma}$ represent the mean and covariance matrix of real and fake feature maps across all video frames, respectively.
Lower FID means the distribution between the feature vectors of real and fake are more similar, hence more realistic fake samples.

\subsection*{B.3. Adversarial Objectives}
\noindent \textbf{Adversarial Losses at Stage \Romannum{1}.}
We employ two discriminators $D^{\text{(\Romannum{1})}}_\text{fr}, D^{\text{(\Romannum{1})}}_\text{sq}$, where each receives Gaussian heatmaps at an individual frame and sequence level, respectively.
$D^{\text{(\Romannum{1})}}_\text{fr}$ plays a role in enhancing the generated keypoints quality better in single frame-level and $D^{\text{(\Romannum{1})}}_\text{sq}$ encourages the generated keypoints sequences to be naturally evolving.
\begin{align}
    &\calL^{\text{(\Romannum{1})}}_\text{adv} = 
    \bbE_{\calH_{t},\hat{\calH}_{t}} \big[\log D^{\text{(\Romannum{1})}}_\text{fr}(\calH_{t}) + \log (1-D^{\text{(\Romannum{1})}}_\text{fr}(\hat{\calH}_{t})) \big] \\
    &+ \bbE_{\calH_{1:T}, \hat{\calH}_{1:T}} \big[\log D^{\text{(\Romannum{1})}}_\text{sq}(\calH_{1:T}) + \log (1-D^{\text{(\Romannum{1})}}_\text{sq}(\hat{\calH}_{1:T}))\big]. \nonumber
\end{align}
In particular, we employ the WGAN-GP~\cite{gulrajani2017improved} loss as the adversarial loss.

\noindent \textbf{Adversarial Losses at Stage \Romannumm{2}.}
We use adversarial losses to achieve both goals: (1) the image discriminator loss for generating realistic frames and (2) the video discriminator loss for retaining motion information, where we use the WGAN-GP~\cite{gulrajani2017improved} loss.
To this end, we use two motion conditional adversarial losses: $D^{\text{(\Romannumm{2})}}_\text{fr}$, $D^{\text{(\Romannumm{2})}}_\text{sq}$ encourages the model to generate realistic frames and retaining motion information. \begin{align}
    \calL^{\text{(\Romannumm{2})}}_\text{adv} &= 
    \bbE_{\bv_{t}, \hat{\bv}_{t}, \calH_{t}}\big[\log D^{\text{(\Romannumm{2})}}_\text{fr}([\bv_{t};\calH_{t}]) + \log (1 - D^{\text{(\Romannumm{2})}}_\text{fr}([\hat{\bv}_{t};\calH_{t}])) \big] \\ 
    &+ \bbE_{\bv_{1:T}, \hat{\bv}_{1:T}, \calH_{1:T}}\big[\log D^{\text{(\Romannumm{2})}}_\text{sq}([\bv_{1:T};\calH_{1:T}]) + \log (1 - D^{\text{(\Romannumm{2})}}_\text{sq}([\hat{\bv}_{1:T};\calH_{1:T}])) \big], \nonumber
\end{align}
where $[\boldsymbol{\cdot} \hspace{0.05cm} ; \hspace{0.05cm} \boldsymbol{\cdot}]$ denotes concatenation. We adopt the WGAN-GP~\cite{gulrajani2017improved} loss similar to stage I.
% make the generated outputs indistinguishable from real videos as well as to reflect the motion information adding Gaussian heatmaps as additional condition to consider.
% Specifically, we adopt two discriminators $D^{img}_{V}, D^{vid}_{V}$ and leverage the WGAN-GP~\cite{gulrajani2017improved} loss.

\begin{figure*}[t!]
    \centering
    \subfigure[RNN-based motion generator]{
        \includegraphics[width=0.99\linewidth]{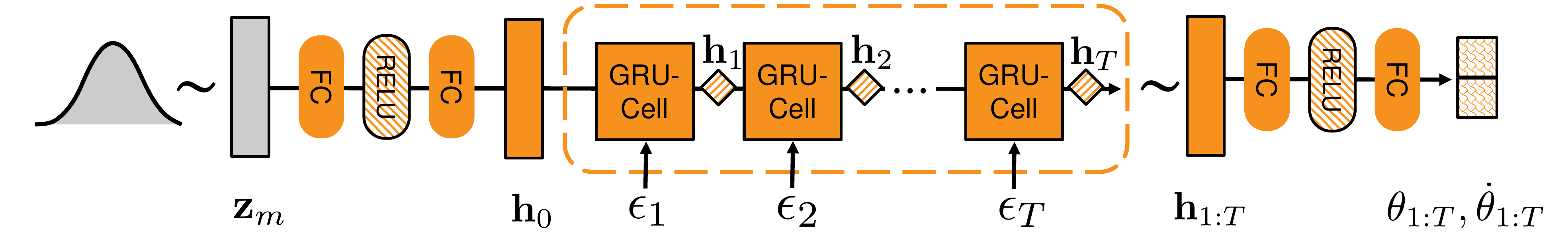}
    }
    \subfigure[Neural ODE-based motion generator]{
        \includegraphics[width=0.99\linewidth]{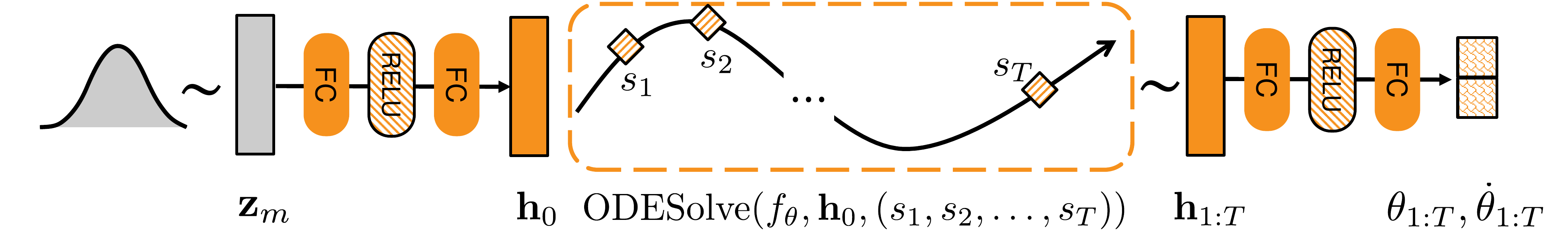}
    }
    \caption{Comparison of RNN and neural ODE-based architectures for pendulum experiment. 
    The only difference lies in the intermediate neural networks that generates the hidden states $\bh_{1:T}$. For the RNN input, a Gaussian noise $\epsilon_{t}$ is used for each timesteps $t=1,2,\ldots T$.}
    \label{fig:pendulum_arch}
\end{figure*}

\subsection*{B.4. Details of Pendulum Experiments}
\noindent \textbf{Dataset Generation.}
\enskip
We used 1,000 simulated pendulum trajectories with fixed gravity force $g$ of 9.81 m/s$^2$. The damping factor $B$, length of pendulum $L$, and mass of bob $M$ were each sampled from the Gaussian distribution with respective means of 0.2, 1.0, 1.0 and unit variance. 
For trajectory simulation, Runge-Kutta was used to solve the pendulum equation (first order differential equations) described below:

\vspace{-0.4cm}
\begin{align*}
        \begin{bmatrix}
        \big( \frac{d\theta}{dt} \big) \\
        \big( \frac{d\dot{\theta}}{dt} \big)
        \end{bmatrix}
        =\begin{bmatrix}
        \dot{\theta} \\ 
        -\big( \frac{g}{L} \big) \cdot \sin(\theta) - \big( \frac{B}{M} \big) \cdot \dot{\theta}
        \end{bmatrix},
\end{align*}

\begin{wrapfigure}{r}{0.4\textwidth}
    \centering
    % \vspace{-1.0cm}
    \includegraphics[width=0.4\textwidth]{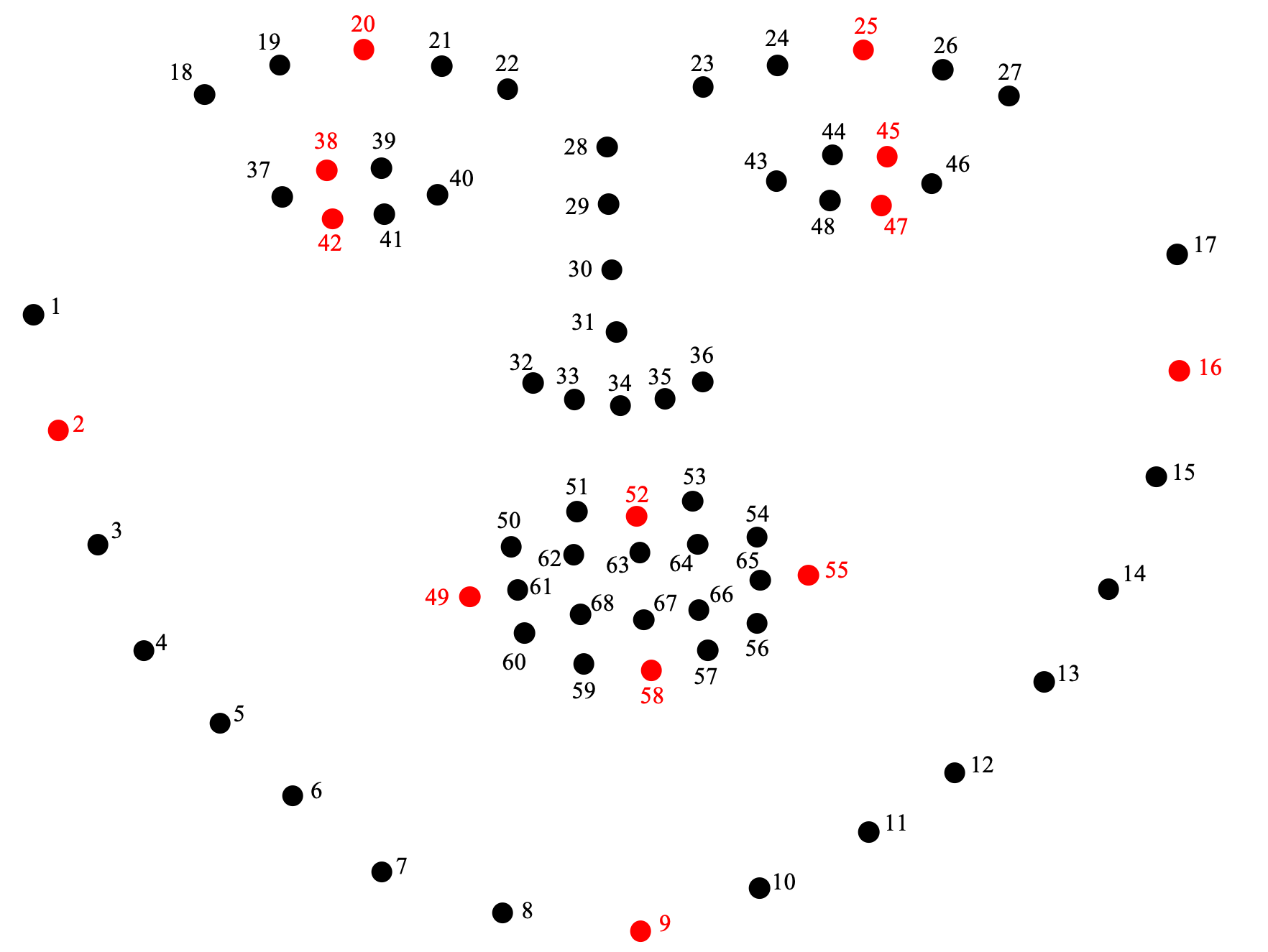}
    \caption{The 68 2D landmarks used by facial landmarks detector model~\cite{bulat2017far}. We use 13 keypoints among them (\textcolor{red}{red}).}
    \vspace{-1.0cm}
    \label{fig:facekeypoints}
\end{wrapfigure}

\noindent where $dt$ is a time unit fixed to 0.1 during data generation.

\noindent \textbf{Comparison of RNN and ODE Motion Generator.}
\enskip
Fig.~\ref{fig:pendulum_arch} shows the detailed architectures of RNN and neural ODE-based pendulum dynamics generator. Both generators are trained using 1,000 simulated pendulum trajectories. After training, the generators should mimic the plausible dynamics of a pendulum by estimating two physical variables $\theta$ and $\dot{\theta}$.

\subsection*{B.5. Dataset Preprocessing}

Fig.~\ref{fig:facekeypoints} shows the chosen facial keypoints (2, 9, 16, 20, 25, 38, 42, 45, 47, 49, 52, 55, 58th) for \textit{MUG}~\cite{aifanti2010mug} and \textit{UvA-NEMO}~\cite{dibekliouglu2012you}.
The facial keypoint selections refer to previous work~\cite{jang2018video}.
We crop and resize to $64 \times 64$ pixels for both datasets.

\begin{figure*}[t!]
    \centering
    \subfigure[Recurrent neural network]{
        \includegraphics[width=0.99\linewidth]{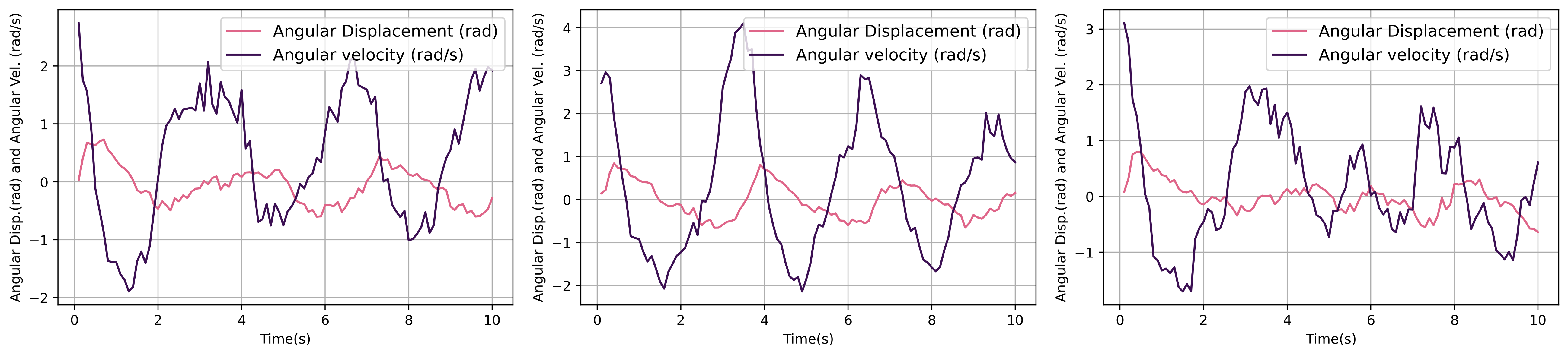}
    }
    \subfigure[Neural ODE]{
        \includegraphics[width=0.99\linewidth]{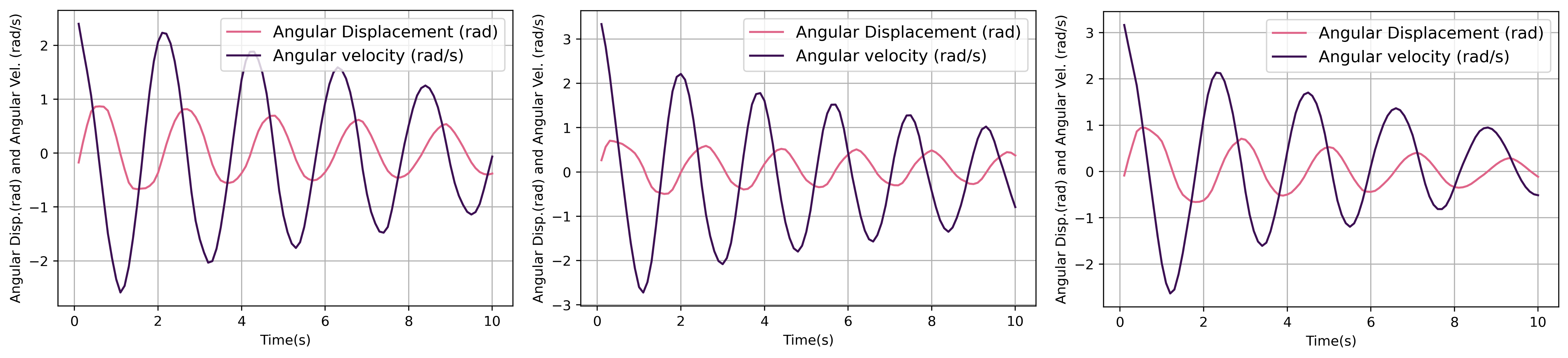}
    }
    \caption{Visualization of generated $\theta, \dot{\theta}$ using two different motoin generators employing (a) RNN and (b) Neural ODE}
    \label{fig:pendulum_examples}
    \vspace{-0.5cm}
\end{figure*}

\section*{\textsc{C. Additional Experiments}}
\label{2_qualitative}

In the following section, we provide additional experiments as follows:
\begin{itemize}
    \item Analysis on the dynamics of each video dataset.
    \item Qualitative examples from pendulum experiments using both recurrent neural network and neural ODE.
    \item Qualitative examples of sampled keypoints sequence from our motion generator.
    \item Qualitative comparison results against unconditional video generation baselines using facial expression datasets.
    \item Qualitative results of appearance and motion manipulation on \textit{UvA-NEMO} and \textit{MUG}.
    \item Qualitative and quantitative evaluations on dynamic frame rate.
\end{itemize}

\begin{figure}[h!]
    \centering
    \subfigure[Weizmann Action]{
        % \label{fig:a}
        \includegraphics[width=\linewidth]{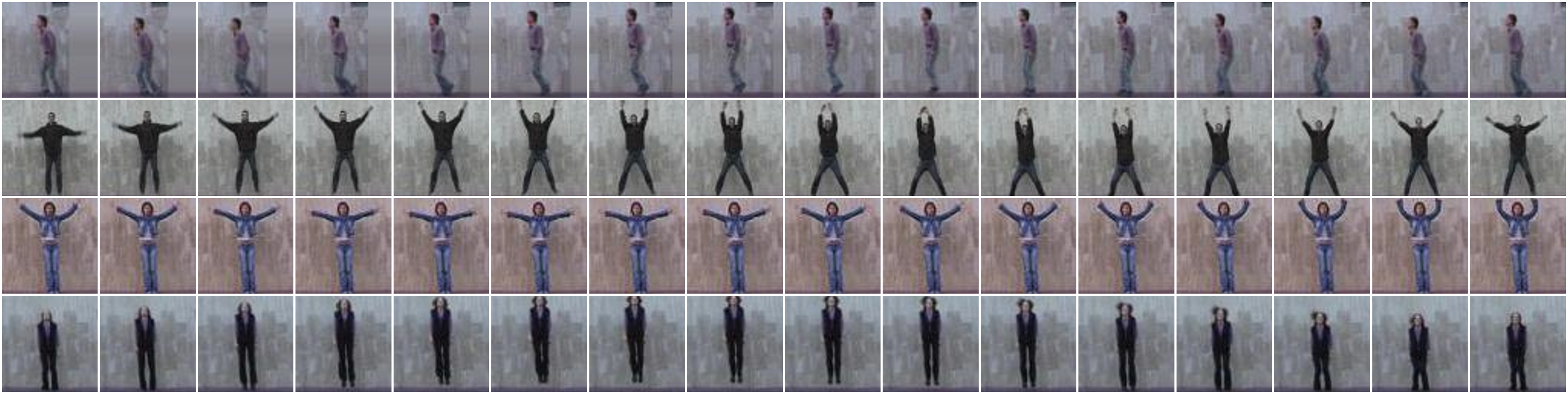}}
    \subfigure[KTH Action]{
        % \label{fig:b}
        \includegraphics[width=\linewidth]{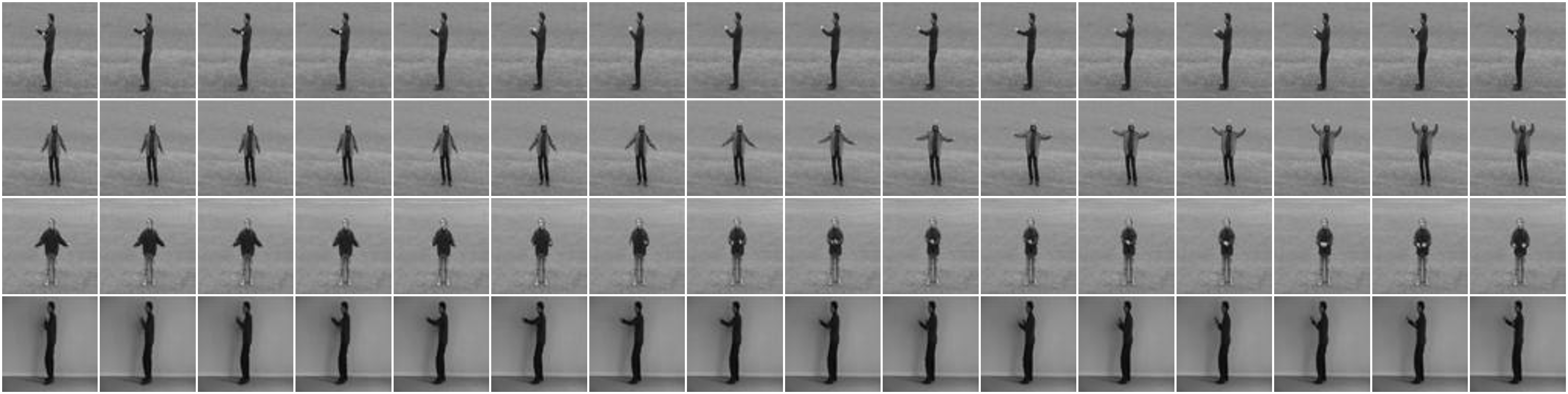}}
    \subfigure[MUG]{
        % \label{fig:c}
        \includegraphics[width=\linewidth]{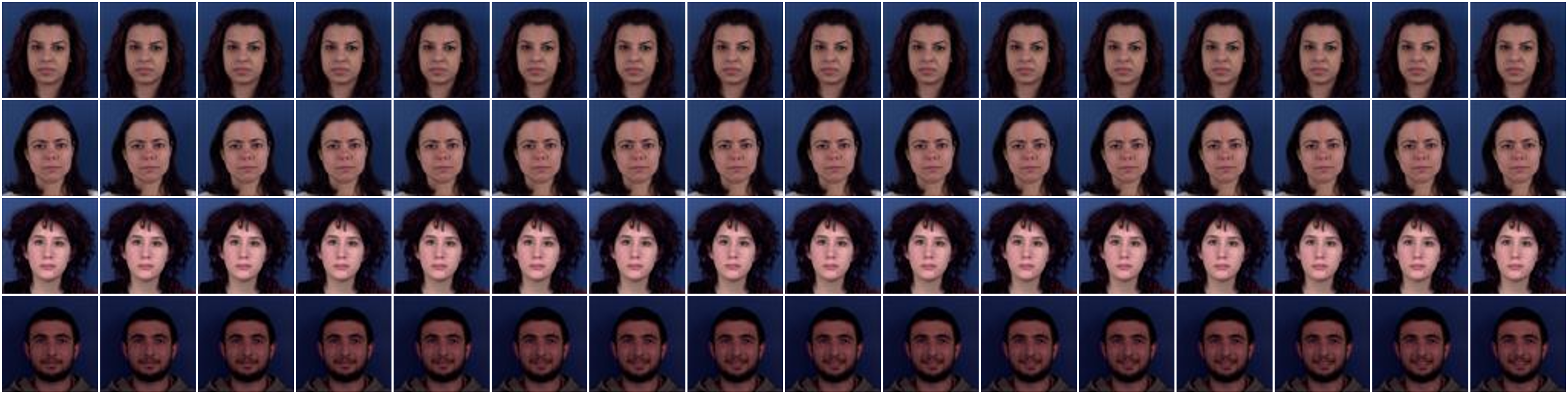}}
    \subfigure[UvA-NEMO]{
        % \label{fig:d}
        \includegraphics[width=\linewidth]{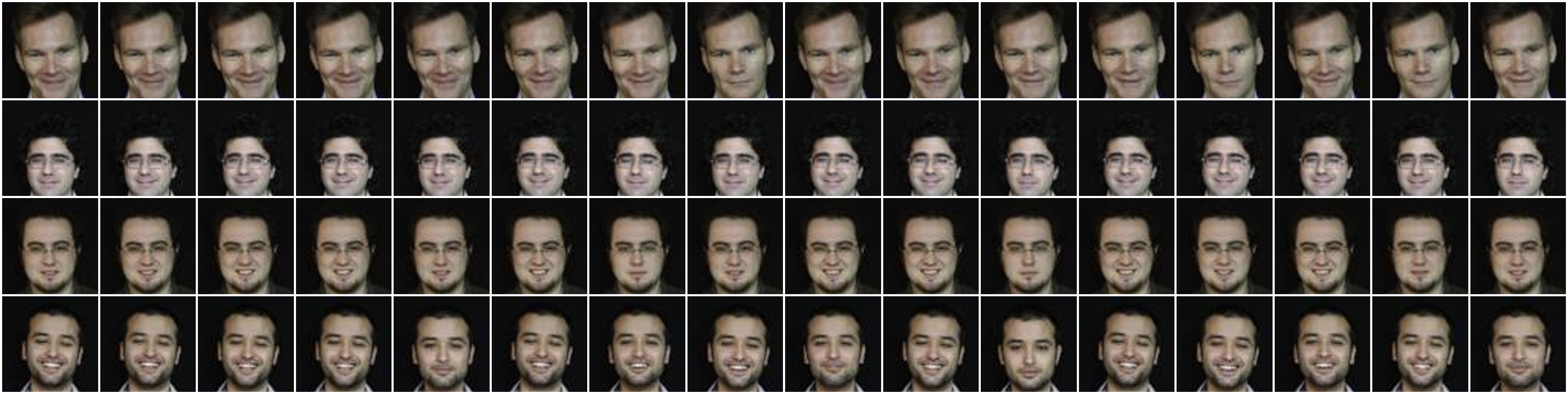}}
    \caption{Video samples from each dataset.}
    \label{fig:dataset}
    % \vspace{-0.5cm}
\end{figure}

\noindent \textbf{Analysis on Dynamics of Datasets.}
To analyze the dynamics of each dataset, we includes the video examples of the four training datasets :\textit{Weizmann Action}, \textit{KTH Action}, \textit{MUG}, and \textit{UvA-NEMO}. 
Due to different frame rates and velocity of human motions in each dataset, the degree of the movement between the adjacent frames varies as shown in Fig.~\ref{fig:dataset}.
We find out that Human action datasets (\textit{i.e.}, \textit{Weizmannn Action} and \textit{KTH Action}) have relatively large movements compared to facial expression datasets (\textit{i.e.} \textit{MUG} and \textit{UvA-NEMO}).
As a result, our model and baselines display more significant movements in the human action videos than in the facial expression videos.

\noindent \textbf{Examples of Pendulum Experiment.}
\enskip
Fig.~\ref{fig:pendulum_examples} shows additional examples of the generated pendulum dynamics from the RNN and the neural ODE based motion generators (See Fig.~\ref{fig:pendulum_arch} for detailed architecture).
As shown in Fig.~\ref{fig:pendulum_examples}, the  RNN-based motion generator fails to generate smooth dynamics (a). In contrast, our ODE-based motion generator successfully simulates the pendulum dynamics (b).

\begin{figure*}[h!]
    \centering
    \includegraphics[width=\linewidth]{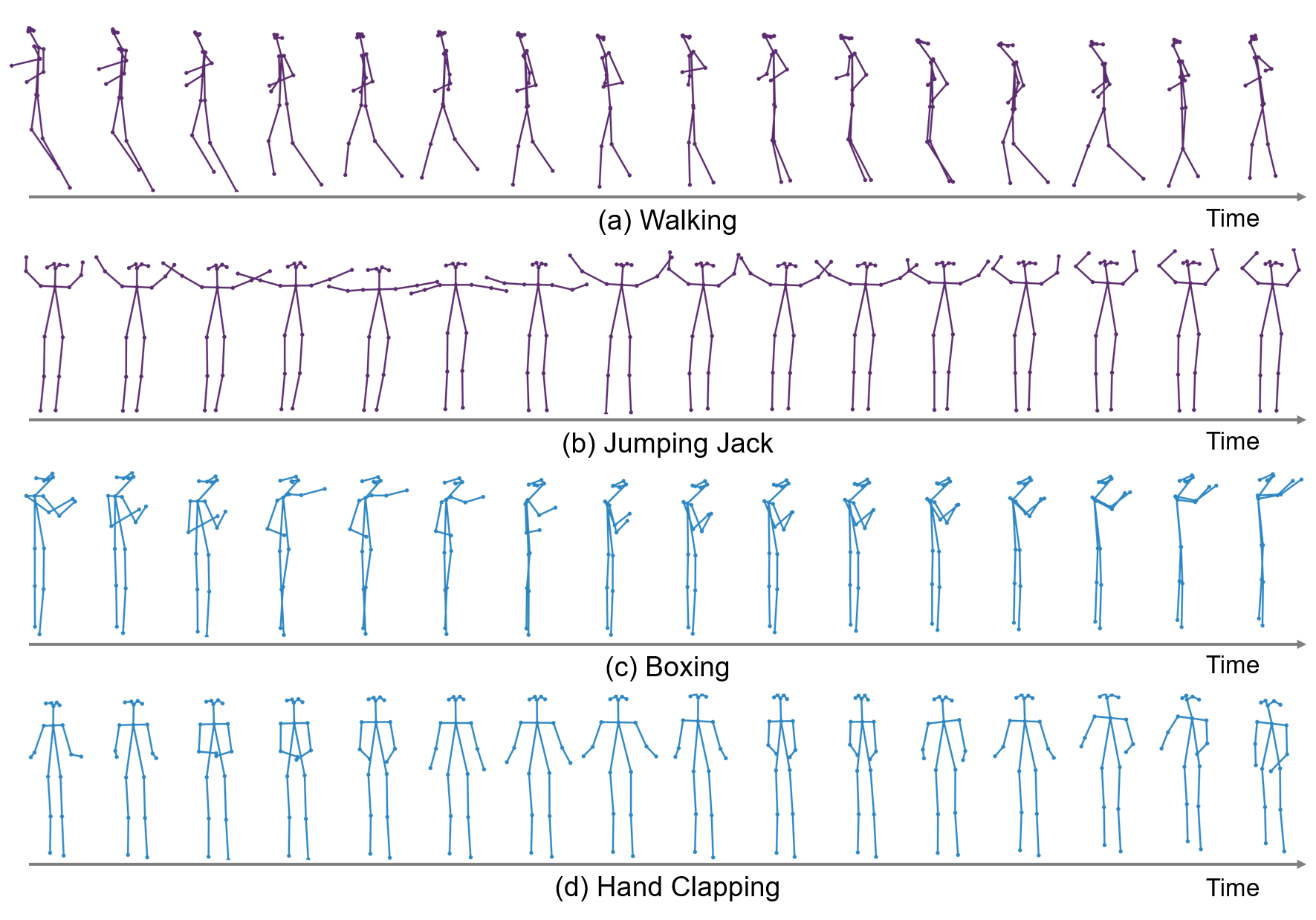}
    \caption{Examples of generated keypoints sequences from our motion generator.}
    \label{fig:supp-pose}
\end{figure*}

\noindent \textbf{Examples of Sampled Motion.}
\enskip
Fig.~\ref{fig:supp-pose} illustrates the examples of the generated keypoints sequences from our motion generator trained on \textit{Weizmann Action} (\textcolor{purple}{purple}) or \textit{KTH Action} dataset (\textcolor{blue}{blue}).
Through these examples, we demonstrate that our motion generator has the capability in producing diverse and fluid motions as a form of a sequence of evolving keypoints.

\begin{figure*}[h!]
    \centering
    \includegraphics[width=\linewidth]{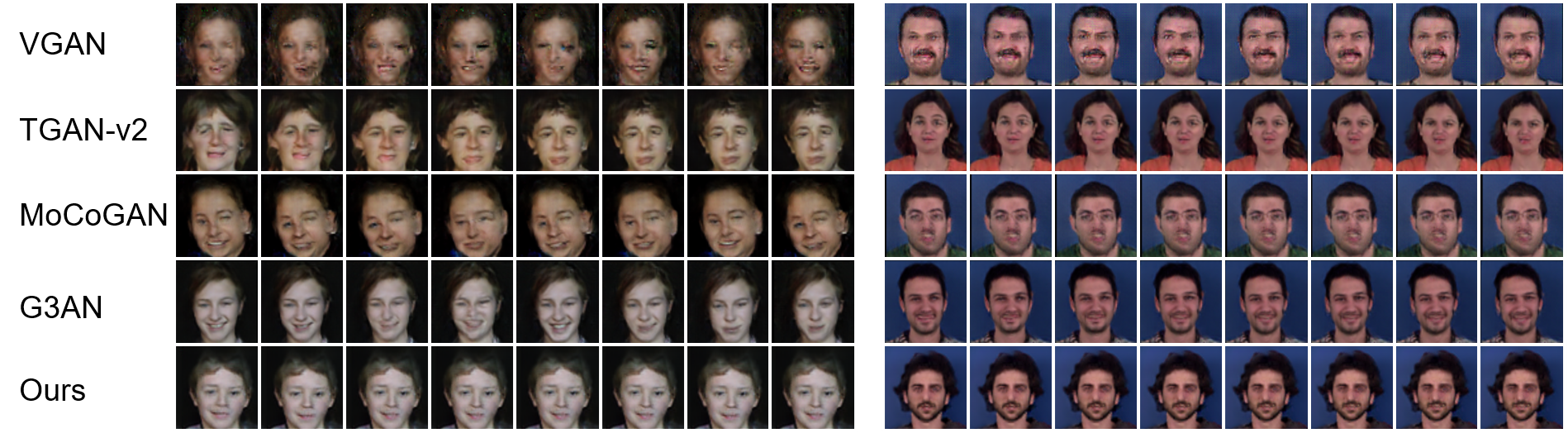}
    \vspace{-0.5cm}
    \caption{Qualitative comparison with baselines on facial expression datasets.}
    \label{fig:supp-qualitative baseline}
    \vspace{-0.2cm}
\end{figure*}

\noindent \textbf{Comparison with baselines using Facial Expression Datasets.}
\enskip
Fig.~\ref{fig:supp-qualitative baseline} presents qualitative comparison of \model with VGAN~\cite{vondrick2016generating}, TGAN-v2~\cite{saitotrain}, MoCoGAN~\cite{tulyakov2018mocogan} and G$^{3}$AN~\cite{wang2020g3an} on two facial expression datasets: \textit{UvA-NEMO} and \textit{MUG}.
\model shows competitive results compared to the baseline models while most generated videos contain relatively marginal movement than human action cases due to the restricted movement characteristics of facial datasets.

% Fig.~\ref{fig:supp-baseline} shows qualitative comparison of \model with VGAN~\cite{vondrick2016generating}, TGAN-v2~\cite{saitotrain}, MoCoGAN~\cite{tulyakov2018mocogan} and G$^{3}$AN~\cite{wang2020g3an} on four different datasets. 
% We followed the hyperparameters as presented in the papers. 

\begin{figure*}[t!]
    \centering
    \includegraphics[width=\linewidth]{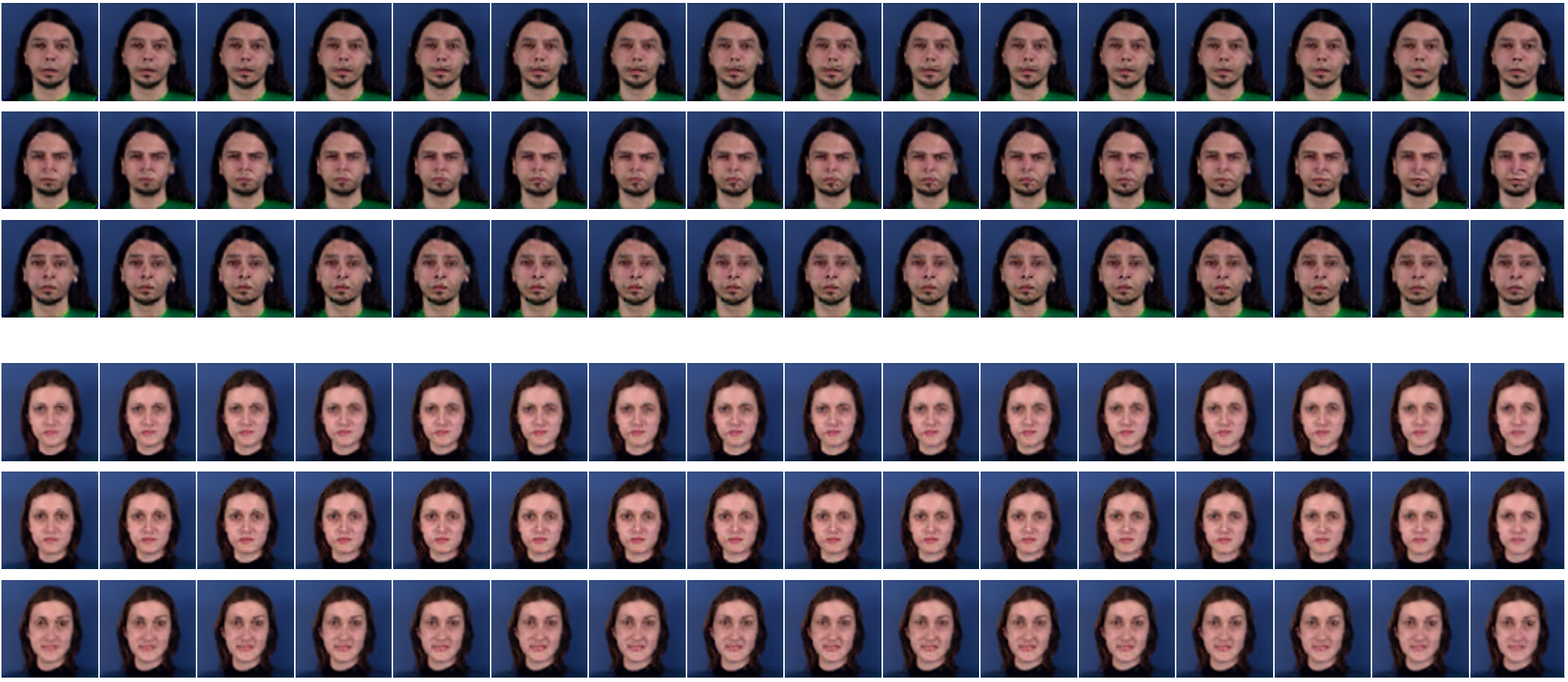}
    \caption{Qualitative results of our method on the \textit{MUG} dataset. Every three row has the same $\bz_a$ (\textit{i.e.} appearance noise vector) with different $\bz_m$ (\textit{i.e.} motion noise vector).}
    \label{fig:mug_fix_appearance}
\end{figure*}

\begin{figure*}[t!]
    \centering
    \includegraphics[width=\linewidth]{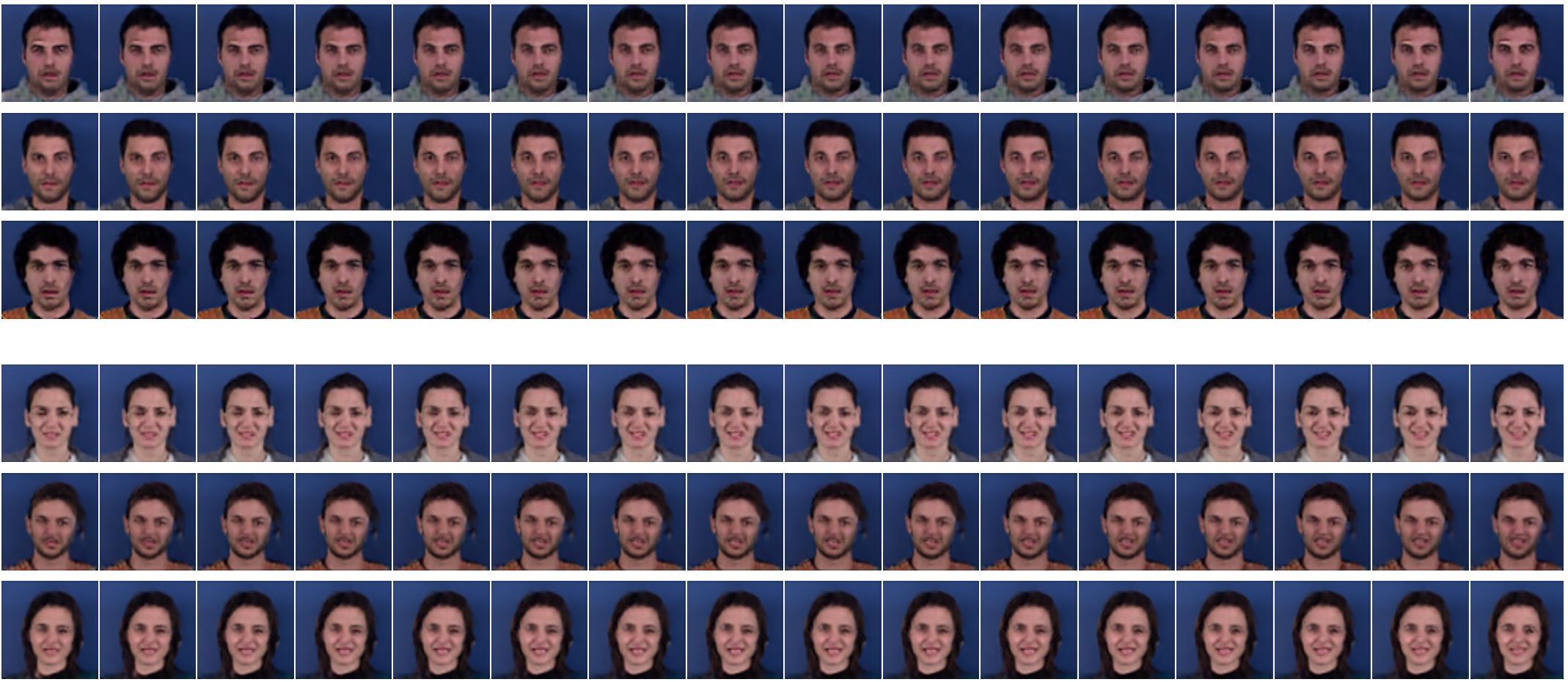}
    \caption{Qualitative results of our method on the \textit{MUG} dataset. Every three row has same $\bz_m$ (\textit{i.e.} motion noise vector) with different $\bz_a$ (\textit{i.e.} appearance noise vector).}
    \label{fig:mug_fix_motion}
\end{figure*}

\begin{figure*}[t!]
    \centering
    \includegraphics[width=\linewidth]{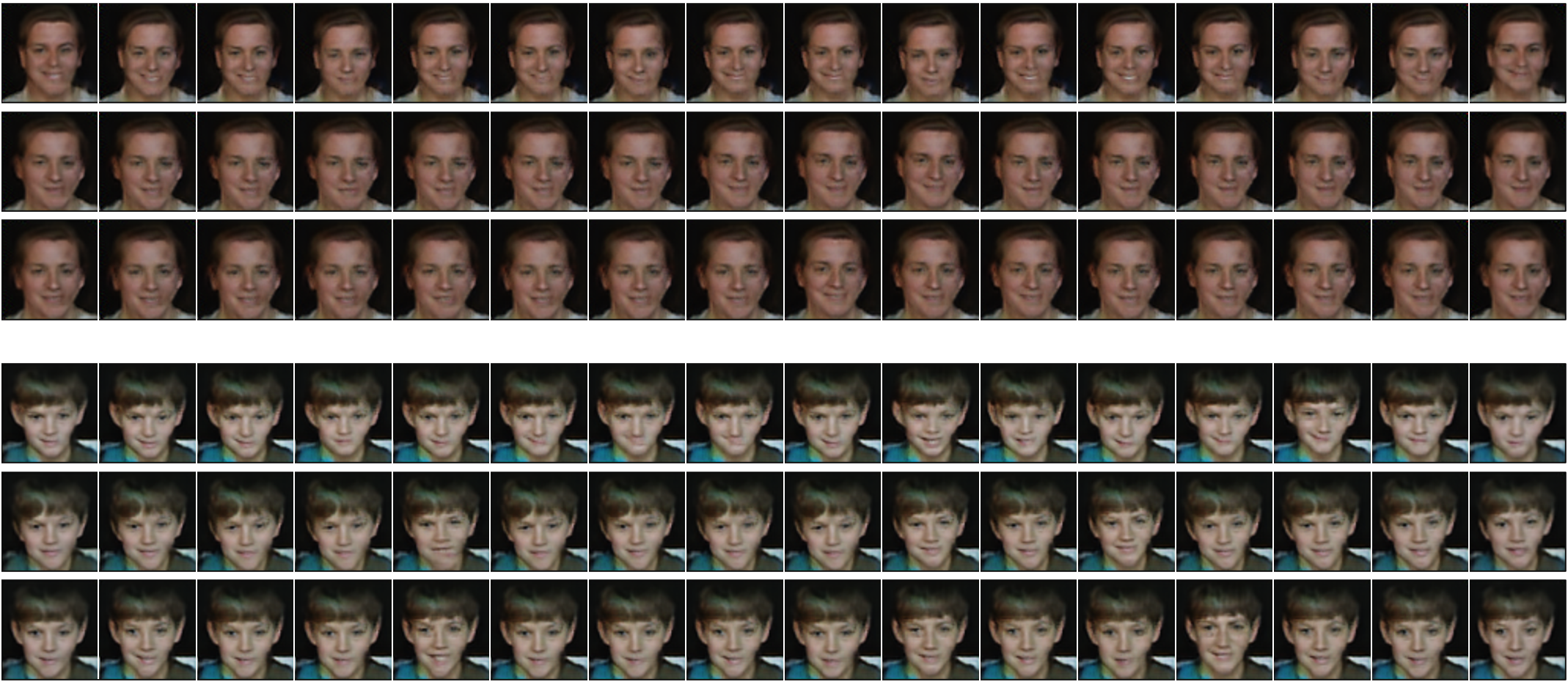}
    \caption{Qualitative results of our method on the \textit{UvA-NEMO} dataset. Every three row has same $\bz_a$ (\textit{i.e.} appearance noise vector) with different $\bz_m$ (\textit{i.e.} motion noise vector).}
    \label{fig:nemo_fix_appearance}
\end{figure*}

\begin{figure*}[t!]
    \centering
    \includegraphics[width=\linewidth]{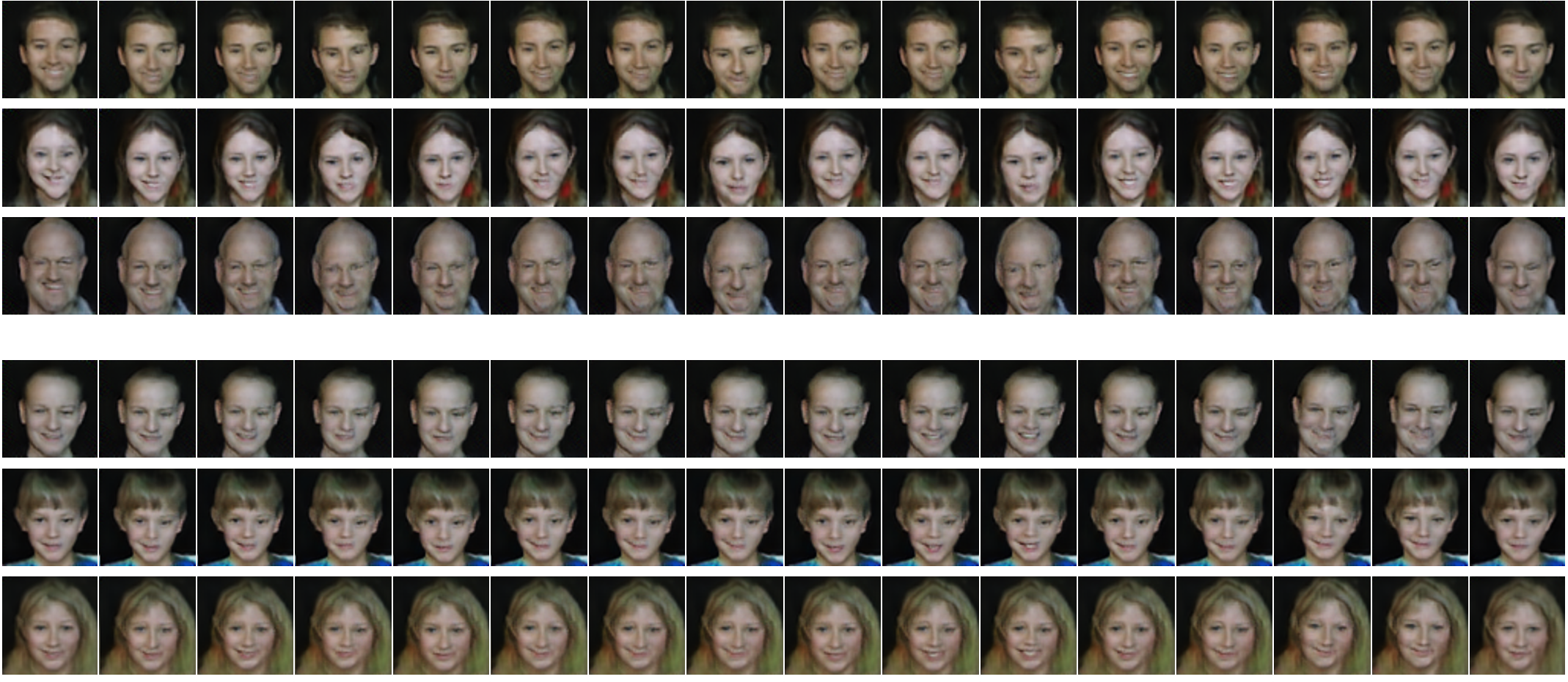}
    \caption{Qualitative results of our method on the \textit{UvA-NEMO} dataset. Every three row has same $\bz_m$ (\textit{i.e.} motion noise vector) with different $\bz_a$ (\textit{i.e.} appearance noise vector).}
    \label{fig:nemo_fix_motion}
\end{figure*}

\noindent \textbf{Examples of Motion and Appearance Manipulation.}
As shown in Fig.~\ref{fig:mug_fix_appearance}-\ref{fig:nemo_fix_motion}, we further demonstrate the ability of \model to separately learn the appearance and motion representations.
We show the examples of manipulating appearance and motion noise vectors in our video generator trained on \textit{MUG} and \textit{UvA-NEMO} dataset.

\begin{figure}[t!]
    \centering
    \subfigure[Ours]{
        % \label{fig:a}
        \includegraphics[width=0.49\textwidth]{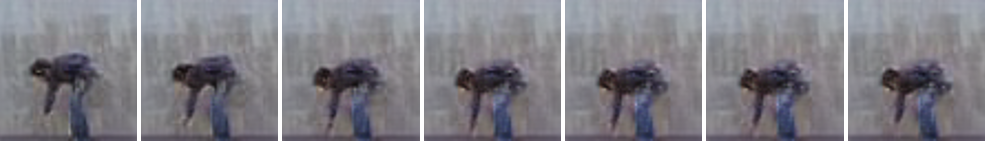}}
    \subfigure[MoCoGAN]{
        % \label{fig:b}
        \includegraphics[width=0.49\textwidth]{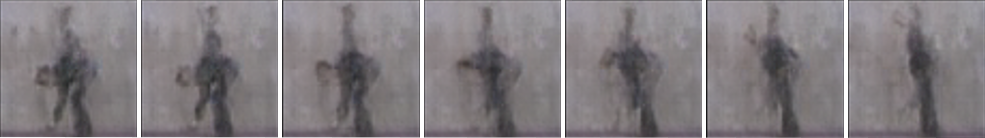}}
    \caption{20-FPS videos generated by two models.}
    \label{exp:continuous_baseline}
    % \vspace{-0.5cm}
\end{figure}

\noindent \textbf{Qualitative Evaluation on Dynamic Frame Rate.}
\enskip
One major strength of \model which makes it distinguished with other baselines is its ability to generate videos in continuous-time domain $\calT$.
In other words, \model can generate realistic videos in arbitrary frame rates regardless of the frame rate of the training samples, by simply integrating (\textit{i.e.} generating keypoints sequences) over the desired timesteps.
(\textit{e.g.,} \model can generate video frames at 0.17 or 0.81 second by training with 1.0-second time intervals).
In other words, \model can flexibly generate a video at an arbitrary frame rate other than the one used at training (\textit{e.g.,} generating video frames at 0.17 or 0.81 second).
In this experiment, we compare the quality of Weizmann Action videos generated in various frame rates ($\{16, 18, 20, 22, 24\}$ FPS) by \model and a baseline (MoCoGAN) both trained at 16 FPS.
While \model learns motion dynamics based on fixed timesteps $\{s_{1},s_{2},\ldots,s_{T}\} \subset \calT, \text{ where }0<s_{1}<s_{2}<\cdots<s_{T}$ during training, it can generate videos in arbitrary frame rates by simply integrating over the desired timesteps during inference.
In this experiment, we compare the quality of videos generated in various frame rates, both by \model and a baseline, namely MoCoGAN.
We define A continuous-time video generation problem aims to generate video frames $\hat{\bv}_{1:L}$ for another set of timesteps $\calM \equiv \{m_{1},m_{2},\ldots,m_{L}\} \subset \calT$.
As a matter of fact, this task can reduces to discrete-time video generation problem as previously tackled if we set $\calM = \calS$.
Furthermore, We evaluate the video quality given the arbitrary length query timesteps $\calM$ to demonstrate the capability on synthesizing a video at continuous-time domain.
As MoCoGAN produces motion codes via its RNN component only for each 1/16 second, we linearly interpolate two adjacent motion codes from the RNN to obtain the motion representation at unseen timesteps.
\footnote{Unlike \model and MoCoGAN, other baselines cannot generate or interpolate motion features at arbitrary timesteps.}
\footnote{We take MoCoGAN as a representative RNN-based model for generating motion features at arbitrary timesteps. Note that G$^3$ AN cannot generate or interpolate motion features at arbitrary timesteps.}
Fig.~\ref{exp:continuous_baseline} shows a comparison between synthetic videos generated by MoCoGAN and \model at 20 FPS.
While MoCoGAN generates frames that looks like a mixture of multiple images, \model generates distinct frames, showing the advantage of the ODE-based approach in handling continuous time.
These results demonstrate that simply interpolating different motion representations to generate in-between frames yields sub-optimal outcomes (\textit{i.e.} mixed frames at the pixel level).

\begin{figure}[t!]
    \centering
    \includegraphics[width=0.7\linewidth]{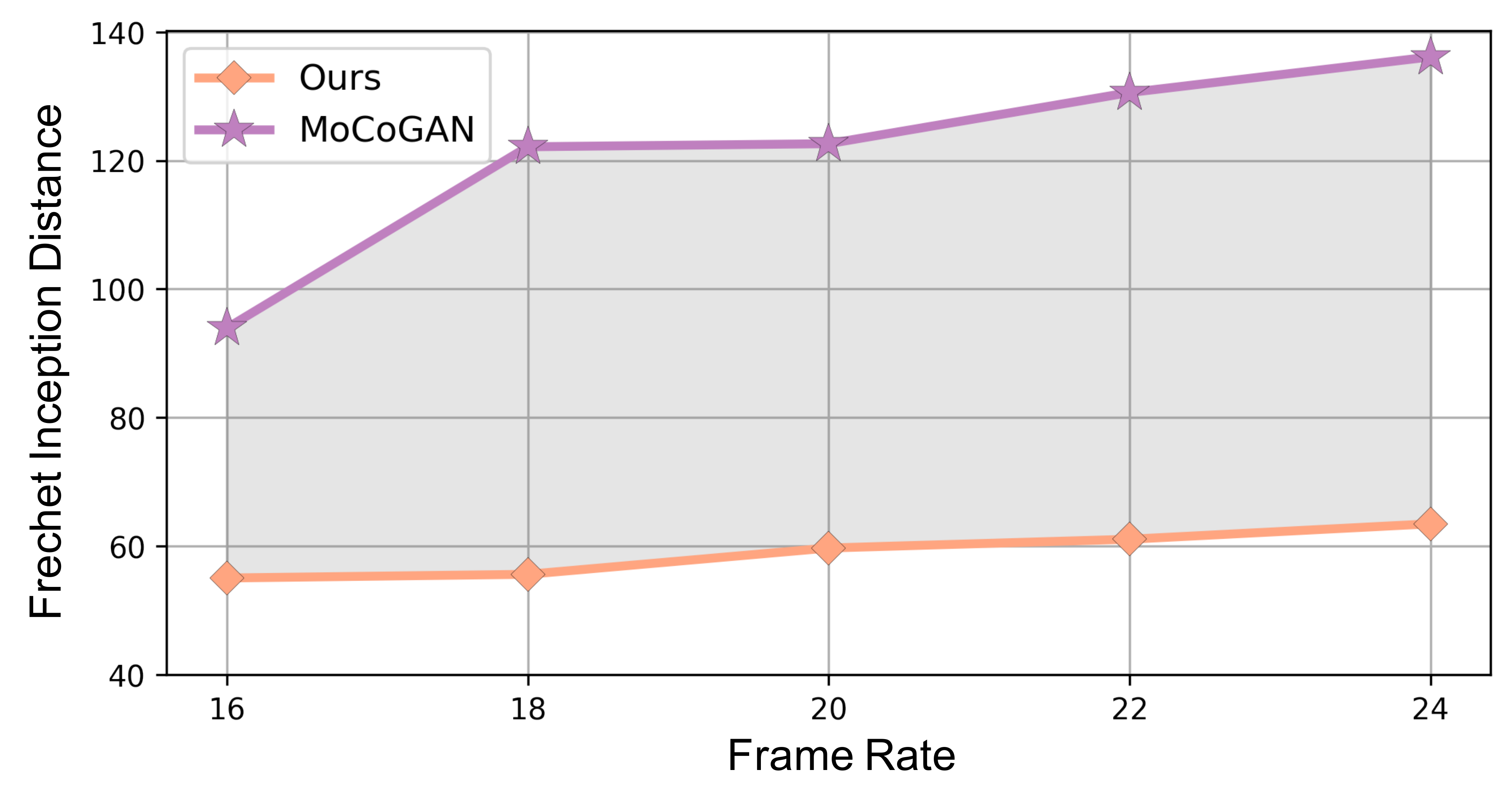}
    \vspace{-0.5cm}
    \caption{Video FID scores with varied frame rates on Weizmann Action. \model consistently outperforms MoCoGAN, showing marginal performance drop to generate at higher frame rate.}
    % \vspace{-0.5cm}
    \label{fig:continuous_fid}
\end{figure}

\noindent \textbf{Quantitative Evaluation on Dynamic Frame Rate.}
Fig.~\ref{fig:continuous_fid} shows the video FID scores of the videos generated at increasing frame rates (from 16 to 24) using the two models traine at 16 FPS.
Based on the the monotonic increase of video FID scores by both models, it is evident that the denser the frame rate, the more challenging the video generation becomes.
However, the steeper slope of MoCoGAN than \model clearly indicates that \model is quite robust to the increasing FPS.
In other words, \model is more comfortable than MoCoGAN in generating videos in continuous-time domain, even at a higher frame rate than the training frame rate.

% \noindent \textbf{Description about the Provided Video}
% \enskip
% Additionally, we include a video to show our results more vividly. 
% The video contains (i) simulated movement of bobs from RNN and neural ODE-based motion generator, (ii) video clips of illustrating successive movement of sampled keypoints sequence, and (iii) video samples moving at multiple velocities (\textit{i.e,} from fast to slow motion).

%%%%%%%%%%%%%%%%%%%%%%%%%%%%%%%%%%%%%%%%%%%%%%%%%%%%%%%%%%%%%%%%%%%%%%%%%%%%%%%%%%%%%%%%%%%%%%%%%%

\clearpage

\end{document}